\documentclass[sigconf]{acmart}

\AtBeginDocument{%
  }

\copyrightyear{2025} 
\acmYear{2025} 
\setcopyright{acmlicensed}
\acmConference[KDD '25]{Proceedings of the 31st ACM SIGKDD Conference on Knowledge Discovery and Data Mining V.2}{August 3--7, 2025}{Toronto, ON, Canada}
\acmBooktitle{Proceedings of the 31st ACM SIGKDD Conference on Knowledge Discovery and Data Mining V.2 (KDD '25), August 3--7, 2025, Toronto, ON, Canada}
\acmISBN{979-8-4007-1454-2/2025/08}
\acmDOI{10.1145/3711896.3737131}

\usepackage{bm}
\usepackage{url}
\usepackage{svg}
\usepackage{tikz}
\usepackage{xcolor}
\usepackage{wrapfig}
\usepackage{amsmath}
\usepackage{caption}
\usepackage{hyperref}
\usepackage{pgfplots}
\usepackage{multirow}
\usepackage{booktabs}
\usepackage{multicol}
\usepackage{graphicx}
\usepackage{placeins}
\usepackage{subfigure}
\usepackage{enumerate}
\usepackage{adjustbox}
\usepackage{algorithm}
\usepackage{algorithmic}
\usepackage{threeparttable}

\begin{document}

\title{SlotPi: Physics-informed Object-centric Reasoning Models}


\author{Jian Li}
\orcid{0000-0002-0685-0861}
\email{lijian2022@ruc.edu.cn}
\affiliation{%
  \institution{Renmin University of China}
  \city{Beijing}
  \country{China}
}

\author{Han Wan}
\orcid{0009-0005-4192-1097}
\email{wanhan2001@ruc.edu.cn}
\affiliation{%
  \institution{Renmin University of China}
  \city{Beijing}
  \country{China}
}

\author{Ning Lin}
\orcid{0000-0001-8883-7005}
\email{linning51400@ruc.edu.cn}
\affiliation{%
  \institution{Renmin University of China}
  \city{Beijing}
  \country{China}
}

\author{Yu-Liang Zhan}
\orcid{0009-0006-4732-6288}
\email{zhanyuliang@ruc.edu.cn}
\affiliation{%
  \institution{Renmin University of China}
  \city{Beijing}
  \country{China}
}

\author{Ruizhi Chengze}
\orcid{0009-0002-5736-0165}
\email{chengzeruizhi@huawei.com}
\affiliation{%
  \institution{Huawei Technologies Ltd.}
  \city{Shanghai}
  \country{China}
}

\author{Haining Wang}
\orcid{0000-0002-7412-3797}
\email{wanghaining5@hisilicon.com}
\affiliation{%
  \institution{Huawei Technologies Ltd.}
  \city{Shenzhen}
  \country{China}
}

\author{Yi Zhang}
\orcid{0000-0003-3487-7073}
\email{zhangyi430@huawei.com}
\affiliation{%
  \institution{Huawei Technologies Ltd.}
  \city{Shanghai}
  \country{China}
}

\author{Hongsheng Liu}
\orcid{0000-0003-0509-7967}
\email{liuhongsheng4@huawei.com}
\affiliation{%
  \institution{Huawei Technologies Ltd.}
  \city{Shanghai}
  \country{China}
}

\author{Zidong Wang}
\orcid{0009-0007-4524-1384}
\email{wang1@huawei.com}
\affiliation{%
  \institution{Huawei Technologies Ltd.}
  \city{Hangzhou}
  \country{China}
}

\author{Fan Yu}
\orcid{0009-0001-2189-351X}
\email{fan.yu@huawei.com}
\affiliation{%
  \institution{Huawei Technologies Ltd.}
  \city{Hangzhou}
  \country{China}
}

\author{Hao Sun}
\orcid{0000-0002-5145-3259}
\authornote{Corresponding author.}
\email{haosun@ruc.edu.cn}
\affiliation{%
  \institution{Renmin University of China}
  \city{Beijing}
  \country{China}
}

\renewcommand{\shortauthors}{Jian Li et al.}

\begin{abstract}

Understanding and reasoning about dynamics governed by physical laws through visual observation, akin to human capabilities in the real world, poses significant challenges. Currently, object-centric dynamic simulation methods, which emulate human behavior, have achieved notable progress but overlook two critical aspects: 1) the integration of physical knowledge into models. Humans gain physical insights by observing the world and apply this knowledge to accurately reason about various dynamic scenarios; 2) the validation of model adaptability across diverse scenarios. Real-world dynamics, especially those involving fluids and objects, demand models that not only capture object interactions but also simulate fluid flow characteristics. To address these gaps, we introduce SlotPi, a slot-based physics-informed object-centric reasoning model. SlotPi integrates a physical module based on Hamiltonian principles with a spatio-temporal prediction module for dynamic forecasting. Our experiments highlight the model's strengths in tasks such as prediction and Visual Question Answering (VQA) on benchmark and fluid datasets. Furthermore, we have created a real-world dataset encompassing object interactions, fluid dynamics, and fluid-object interactions, on which we validated our model's capabilities. The model's robust performance across all datasets underscores its strong adaptability, laying a foundation for developing more advanced world models. We will release our codes and data at \textit{\url{https://github.com/intell-sci-comput/SlotPi}}.

\end{abstract}





\maketitle

\section{Introduction}

\begin{figure*}[t!]
\begin{center}
\centerline{\includegraphics[width=0.96\linewidth]{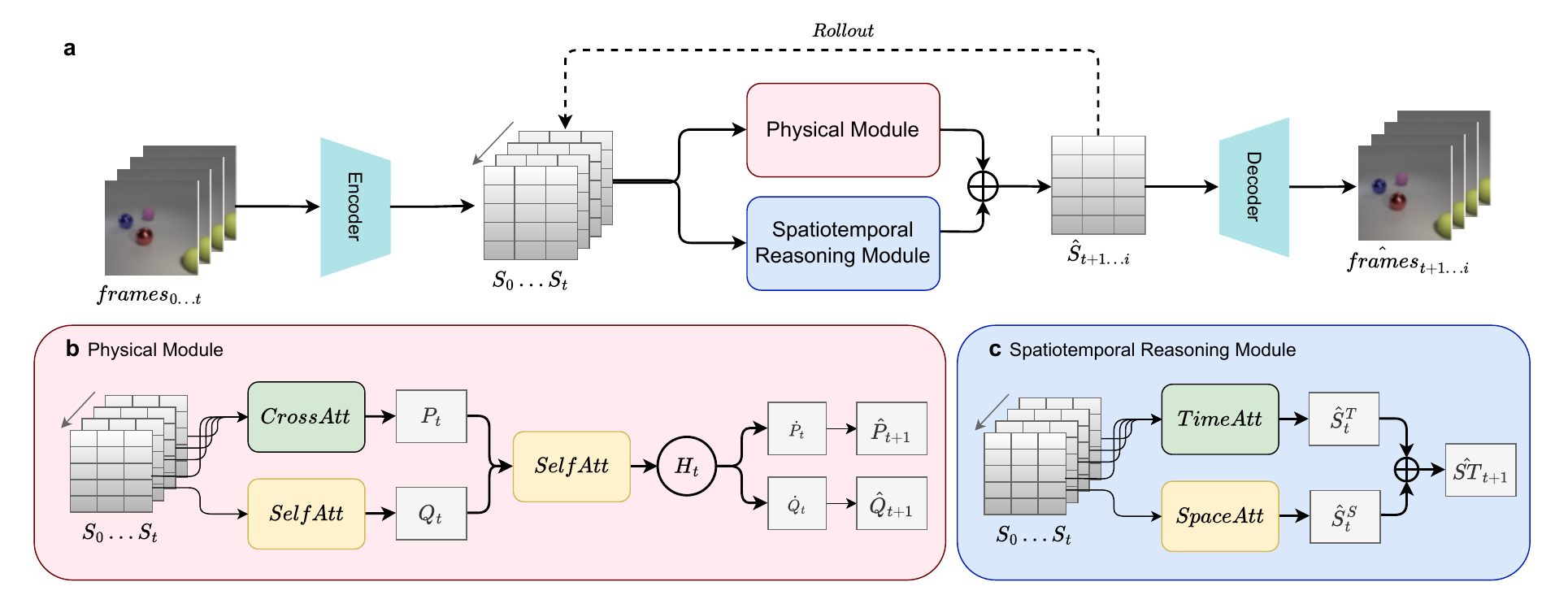}}
\vspace{-10pt}
\caption{(a) Overall framework. SlotPi employs slots extracted by the object-centric model for reasoning and is mainly composed of two modules: a physics module and a spatiotemporal reasoning module. (b) Physics module. This module is designed based on Hamiltonian equations to provide physical constraints for reasoning of SlotPi. (c) Spatiotemporal reasoning module. This module captures and reasons about the parts of the system that cannot be directly  inferred from the physics module.} 
\label{fig1}
\vspace{-10pt}
\end{center}
\end{figure*}

The world is governed by physical laws that dictate the movement, interaction, and transformation of matter. Humans acquire knowledge by perceiving and observing the physical world, enabling a deep understanding of their surroundings—an essential aspect of human intuitive physics \cite{shiffrin1977controlled, lake2017building, kubricht2017intuitive, riochet2018intphys, smith2019promise}. Objects serve as the fundamental units of human perception, cognition, and reasoning. By analyzing interactions between objects and their environment, humans can infer the state of objects and their surroundings. Human intuitive reasoning relies heavily on an inherent understanding of object behavior patterns and physical laws \cite{gopnik1992child}. Cognitive science research shows that the human brain rapidly reasons and makes judgments by constructing internal models of objects and their interrelationships \cite{spelke2007core, baillargeon1987object, clark2013whatever}. These judgments are grounded in a profound understanding of physical laws, such as the principles of mechanics \cite{shiffrin1977controlled}. Therefore, simulating human intuitive physical behavior to construct object-centric models is critical for developing more adaptive artificial intelligence systems that align with human cognitive mechanisms.

Currently, slot-based object-centric learning methods have made notable progress in scene decomposition \cite{kipf2021conditional, elsayed2022savi++, wu2023SlotFormer}. These methods allocate the representations of different objects or backgrounds within a scene to specific slots, making the decomposed slot features more flexible. This flexibility facilitates their use by downstream world models to perform interaction computations, thereby simulating the interactions and dynamics among different entities in the scene more accurately. 

Slot-based predictive world models primarily include SlotFormer \cite{wu2023SlotFormer} and STATM \cite{li2024reasoningenhanced}, among others. SlotFormer is a Transformer-based object-centric dynamics model that builds upon object-centric features. The model performs dynamic simulation by utilizing self-attention mechanisms across all historical slots to compute interactions. Inspired by human spatiotemporal reasoning behavior, STATM applies spatiotemporal attention mechanisms to slots to infer the temporal motion states of objects and calculate spatial interactions between objects, while also integrating both time and space attention results. These slot-based object-centric dynamic simulation methods have demonstrated promising performance.

However, these object-centric dynamic simulation methods  overlook two critical aspects: 1) They do not incorporate physical knowledge into their models. In reality, humans develop common-sense knowledge through daily observations and learning, which aids in better understanding the dynamic world. For example, we observe that moving objects (in the absence of external forces) and flowing water do not abruptly stop, leading to the formation of "common-sense" insights. When reasoning about object motion, we intuitively apply this "common-sense," even if we do not explicitly recognize its connection to the law of conservation of energy. 2) They lack validation of model adaptability across more diverse datasets, especially real-world scenarios. The real world encompasses more intricate interactions, such as those between fluids and objects. In such scenarios, fluids are inherently dynamic, with their motion influencing object movement, while object motion, in turn, affects fluid flow, further impacting the motion of other objects. This complexity demands models capable of not only computing object-object interactions but also understanding and learning the unique characteristics of fluids.

Inspired by the human ability to acquire commonsense physical laws, we propose a slot-based physics-informed object-centric reasoning model, termed \textbf{SlotPi} . The model combines a Hamiltonian-based physical module with a spatiotemporal reasoning module to make accurate dynamic predictions. The physical module follows the structure of Hamilton's equations, utilizing slot-based time and space attention mechanisms to solve for generalized momentum and generalized coordinates. It then uses self-attention to calculate the scene’s energy and update its state. The slot-based spatiotemporal reasoning module is designed to capture and reason about the parts of the system that cannot be directly inferred from the physical module. We evaluated the model’s performance on video prediction tasks using benchmark datasets with complex object interactions and validated its effectiveness in downstream Visual Question Answering (VQA) tasks. Additionally, we tested the model’s dynamic reasoning capabilities on a fluid dataset (without objects) and a real-world dataset involving fluid-object interactions. Experimental results demonstrate that our model excels across benchmark datasets, fluid dataset, and real-world dataset featuring fluid-object interactions. In summary, our contributions are as follows:
\begin{itemize}
\item We propose a novel, generalizable latent Hamiltonian network module that explicitly models temporal dynamics and latent physical interactions for high-level physical reasoning. It supports unified modeling of heterogeneous physical systems (e.g., rigid bodies, fluids, and their interactions) while preserving decoupled representations of internal dynamics.
\item We introduce SlotPi, a new unsupervised object-centric reasoning model that goes beyond the limitations of conventional rigid-body and particle-based systems. SlotPi effectively models complex dynamical systems such as rigid-bodies, fluids and coupled fluid-object interactions, exhibiting strong adaptability and advanced physical reasoning capabilities.
\item We construct a real-world dataset featuring scenarios with higher complexity, such as fluid-object interactions. SlotPi achieves state-of-the-art results on all datasets.
\end{itemize}

\section{Related Work}

\textbf{Object-centric Learning.} Our work relies on upstream slot-based, object-centric learning models applied to video data for slot extraction \cite{kipf2021conditional, elsayed2022savi++, singh2022simple, li2024reasoningenhanced}. The primary goal of object-centric learning is to decompose complex scenes into individual object representations, allowing the model to focus on understanding and reasoning about the distinct entities within an environment. Earlier efforts in this domain include SQAIR \cite{kosiorek2018sequential}, which introduced a sequential method for learning object-centric representations; R-SQAIR \cite{stanic2019r}, an extension that improves upon the original framework with relational reasoning between objects; SCALOR \cite{jiang2019scalor},which introduces a scalable architecture for object-centric learning and others\cite{burgess2019monet, veerapaneni2020entity, weis2020unmasking, bear2020learning, kabra2021simone}. Slot-based models represent a significant strategy in object-centric learning, uniquely identifying each object in a scene as an individual slot that stores specific features and attributes of the object \cite{kumar2020ma, yang2021self, dittadi2021generalization, hassanin2022visual, wu2023SlotFormer}.

\textbf{Object-centric Prediction.} Research into object-centric reasoning methods is crucial for human intelligence and also a key objective in artificial intelligence. Early work related to object-centric prediction includes Interaction Network \cite{battaglia2016interaction}, Neural Physics Engine \cite{chang2016compositional}, Visual Interaction Networks \cite{watters2017visual}, and many models developed based on them \cite{chen2021roots, jusup2022social, meng2022physics, piloto2022intuitive, singh2022simple, driess2023learning}. In recent years, slot-based object-centric predictive models have seen significant advancements \cite{ding2021dynamic, ding2021attention, wu2023SlotFormer}. To gain a deeper understanding of commonsense intuitive physics within artificial intelligence systems, \cite{piloto2022intuitive} has developed a system capable of learning various physical concepts. However, these studies often fail to use physical knowledge to enhance predictions as humans do, nor have they validated their performance in a broader range of scenarios. Our primary research goal is to simulate human intuitive physical behavior and construct a more adaptable object-centric reasoning system that integrates universal physical laws.

\textbf{Fluid Prediction.} Fluid prediction techniques can be broadly classified into two main categories: purely data-driven methods and physics-aware approaches. Data-driven approaches directly learn mappings between function spaces through data, includes Conv-based NN models \cite{stachenfeld2021learned, bar2019learning}, U-Net \cite{gupta2022towards}, ResNet \cite{he2016deep}, DeepONet \cite{lu2021learning}, multiwavelet-based model (MWT) \cite{gupta2021multiwavelet}, Fourier Neural Operator (FNO) \cite{li2020fourier}, graph neural networks \cite{sanchez2020learning, pfaff2020learning}, and Transformer-based models \cite{janny2023eagle, li2024scalable, wu2024transolver}. Physics-aware methods utilize information from partially known fluid dynamics equations(e.g., PINN \cite{raissi2019physics, raissi2020hidden, wang2020towards, tang2023adversarial}, PhyGeoNet \cite{gao2021phygeonet}, PhyCRNet \cite{ren2022phycrnet}, PhySR \cite{ren2023physr}, PeRCNN \cite{rao2022discovering, rao2023encoding}, TiGNN \cite{hernandez2023thermodynamics}, EquNN \cite{wang2020incorporating}, and PDE-Net models \cite{long2018pde, long2019pde} , etc.), such as initial/boundary conditions (I/BCs). These methods integrate partial differential equation (PDE) losses (physical losses) or embed the specific structures of equations (like I/BCs) directly into the network architecture using convolution kernels. Our approach to physical embedding markedly differs from these physics-aware strategies. We do not depend on pre-known fluid equations or incorporate such specifics into our neural network framework. Instead, our method uniquely embeds the universal physical laws of the physical world into the model, is also a data-driven method.

\textbf{Hamiltonian Systems.} The applications of Hamiltonian systems in deep learning primarily include Standard Hamiltonian systems and Generalize Hamiltonian systems \cite{chen2022learning}. Standard Hamiltonian systems require pre-known information, such as the coordinates of the system \cite{howse1995gradient, seung1997minimax, raissi2019physics, greydanus2019hamiltonian, brockman2016openai}. Generalize Hamiltonian systems can autonomously learn these coordinates from data through the model itself \cite{howse1995gradient, course2020weak, lee2021machine, chen2021neural, sosanya2022dissipative}. Although Hamiltonian dynamics models demonstrate significant advantages in simulating complex physical systems, existing models are typically designed only for conservative systems. This design limitation makes them less adaptable to new or unknown types of physical phenomena, especially in systems involving nonconservative forces, such as friction or external driving forces. Therefore, the goal of this paper is to design a highly adaptable system embedded with Hamiltonian theory.

\section{SlotPi: Physics-informed Object-centric Reasoning}

In this paper, we introduce \textbf{SlotPi}, a physics-informed object-centric reasoning model designed to align inference results more closely with physical laws. This model is comprised of two primary components: 1) a physical module and 2) a spatiotemporal reasoning module. The physical module, developed by emulating Hamiltonian equations, provides a robust foundation that adheres to physical laws. The spatiotemporal reasoning module is tasked with capturing and reasoning the parts of system that cannot be directly inferred from the physical module. These modules work in synergy to enhance each other's functionality and effectiveness. The overall framework is shown in Figure~\ref{fig1}.

SlotPi functions through slots, so we first use object-centric methods to extract slot representations from frames. Given $T$ input frames $\{x_0, \dots, x_t\}$, we can get $\{S_0, \dots, S_t\} = f_{OC}(\{x_0, \dots, x_t\})$, where $S_t= \{s_{(0,t)}, \dots, s_{(N,t)}\} \in \mathbb{R}^{N \times D}$ represents the $N$ slots information at time $t$. $f_{OC}$ denotes object-centric models such as SAVi \cite{kipf2021conditional} or STATM-SAVi \cite{li2024reasoningenhanced}.

\subsection{Physical Module}

The physical module is designed based on Hamiltonian equations, leveraging slot data extracted by an object-centric model. 

Initially, a cross-attention mechanism and a self-attention mechanism are used to compute the generalized momentum $P_t$ and generalized coordinates $Q_t$ of slots $S_t$, as shown below:
\begin{subequations}
\label{eq1}
\begin{align}
P_t &= \mathrm{CrossAtt}_P (S_t,\{S_0, \dots, S_{t-1}\}) \in \mathbb{R}^{N \times D}, \\
Q_t &= \mathrm{SelfAtt}_Q (S_t) \in \mathbb{R}^{N \times D} .
\end{align}
\end{subequations}
To simplify the calculations, we use a Corresponding Slot Attention method similar to that in STATM-SAVi \cite{li2024reasoningenhanced} in Equation~(\ref{eq1}a). Specifically, for $ith$ slot $s_{(i,t)}$ of $S_t$, attention is computed using this slot along with its corresponding slots in $\{s_{(i,0)}, \dots, s_{(i,t-1)}\}$.

Using the solved values of coordinates $Q_t$ and momentum $P_t$ for $S_t$, self-attention is applied to compute the total energy $H_t$ of the system as follows:
\begin{equation}
\label{eq2}
SH_t = \mathrm{Linear}(\mathrm{SelfAtt}_H(Q_t,P_t)) \in \mathbb{R}^{N \times 1} .
\end{equation}
Here, $SH_t$ represents the energy values of the $N$ slots. To ensure that 
the total energy value $H_t \in \mathbb{R}^+$ of the entire scene remains positive, a linear layer with a Softplus activation function is added following the results from the self-attention mechanism. By summing up the slot energies $SH_t$ at the current moment, the total energy value $H_t$ of the entire scene can be obtained.

Based on the $H_t$, the rates of change for momentum $P_t$ and coordinates $Q_t$ are derived. Finally, Euler's method is applied to compute the momentum $\widehat{P}_{t+1}$ and coordinates $\widehat{Q}_{t+1}$ for the next moment. The corresponding formulations are given by :
\begin{equation}
\label{eq3}
\dot{Q}_t = \frac{\partial H_t}{\partial P_t} \in \mathbb{R}^{N \times D}, \quad \dot{P}_t = -\frac{\partial H_t}{\partial Q_t} \in \mathbb{R}^{N \times D} ,
\end{equation}
\begin{subequations}
\label{eq4}
\begin{align}
\widehat{Q}_{t+1} &= Q_t + \dot{Q}_t \cdot \Delta t \in \mathbb{R}^{N \times D} ,\\
\widehat{P}_{t+1} &= P_t + \dot{P}_t \cdot \Delta t \in \mathbb{R}^{N \times D} .
\end{align}
\end{subequations}
To prevent excessive energy fluctuations, $\Delta t$ is set to a small value based on the video's frame rate (fps) or simulation time step.

\subsection{Spatiotemporal Reasoning Module}

Considering that the motion of real-world objects involves energy dissipation and that energy changes are typically non-conservative, Hamiltonian physics module alone cannot fully capture the motion of objects within a scene. Therefore, we have integrated a spatiotemporal reasoning module that works in conjunction with the physics module to achieve more accurate predictions of dynamics.

Similar to STATM \cite{li2024reasoningenhanced}, for temporal dynamic reasoning, a cross-attention mechanism is employed. Meanwhile, for computations involving spatial interactions, we utilize a self-attention mechanism that operates on slot representations to compute the relevance between different slots within $S_t$, described as: 
\begin{equation}
\label{eq5}
\widehat{ST}_{t+1} = \mathrm{CrossAtt}_T(S_t,\{S_0, \dots, S_t\})+\mathrm{SelfAtt}_S(S_t) \in \mathbb{R}^{N \times D} ,
\end{equation}
where $\widehat{ST}_{t+1}$ is the output of the spatiotemporal reasoning module.

Ultimately, the overall prediction result $\widehat{S}_{t+1}$ of the model is obtained by the weighted sum of the outputs from the physical module $\widehat{Q}_{t+1}$ and the spatiotemporal reasoning module $\widehat{ST}_{t+1}$: 
\begin{equation}
\label{eq6}
\widehat{S}_{t+1} = \lambda \widehat{Q}_{t+1} + \widehat{ST}_{t+1} \in \mathbb{R}^{N \times D} .
\end{equation}
Aside from the ablation studies, $\lambda$ for all other experiments is set to 1. In the ablation study section, we will investigate the impact of different $\lambda$ values on the model's performance to better understand the contribution of each module to the overall model.

\subsection{Training}

Similar to previous work \cite{kipf2021conditional, li2024reasoningenhanced}, in order to enhance the model's long-term prediction quality, we train the model by minimizing the $\ell_2$ loss between the predicted slots and ground-truth slots, simulating the autoregressive generation process during testing. The slot reconstruction loss function $\mathcal{L}_S$ is defined as:  
\begin{equation}
\label{eq7}
\mathcal{L}_S = \frac{1}{M \cdot N} \sum_{m=1}^{M} \sum_{i=0}^{N} \left\| \hat{s}_{(i,t+m)} - s_{(i,t+m)} \right\|^2,
\end{equation}
where $M$ represents the number of rollout steps during the training phase. When using SAVi and STATM-SAVi as the object-centric models for slot extraction, we also introduce an image reconstruction loss $\mathcal{L}_I$ (also $\ell_2$) to jointly train the model. Therefore, the final loss function $\mathcal{L}$ is defined as:
\begin{equation}
\label{eq8}
\mathcal{L} = \mathcal{L}_S + \mathcal{L}_I .
\end{equation}

Additionally, when training the model on the fluid dataset, we use only MSE loss for image reconstruction.

\section{Experiments}

Our experiments primarily comprise four parts: 1) prediction and VQA on benchmark datasets, 2) prediction on fluid dataset, 3)  prediction on real-world dataset, and 4) ablation studies. The first part aims to demonstrate the robust performance of SlotPi in prediction and downstream VQA tasks on benchmark datasets. The second part of our study is designed to validate the model's prediction capabilities on fluid dataset, thereby verifying its adaptability. The third part is aimed at validating the model's performance on real-world dataset that encompass phenomena such as fluid and object interactions.  
The ablation studies focus on assessing the impact of different components of the model or varying experimental settings on the model's performance.

\textbf{Baselines.} In Section~\ref{exp-41}, we primarily compare with SlotFormer \cite{wu2023SlotFormer}, SAVi-Dyn, and STATM \cite{li2024reasoningenhanced}. SlotFormer \cite{wu2023SlotFormer} is a transformer-based framework for object-centric visual simulation. It utilizes slots extracted by upstream object-centric models to train a slot-based transformer encoder for prediction tasks. Additionally, SlotFormer leverages rollout simulation results to train Aloe \cite{ding2021attention} for Visual Question Answering (VQA) tasks. STATM \cite{li2024reasoningenhanced} applies spatiotemporal attention mechanisms to slots to infer the temporal motion states of objects and calculate spatial interactions between objects. When performing VQA tasks, STATM also leverages rollout simulation results to train Aloe, similar to SlotFormer. In the VQA task on Physion \cite{bear2021physion}, we also performed a simple comparison with scene centric methods such as VDT \cite{lu2023vdt} and PRIN.  For more description, please refer to Appendix Section~\ref{append-baseline}.

\textbf{Metrics.} To evaluate the prediction outcomes (excluding fluids), we utilize FG-ARI and FG-mIoU, which are similar to the Adjusted Rand Index (ARI) \cite{rand1971objective, hubert1985comparing} and the mean Intersection over Union (mIoU), except that they exclude background pixels in the calculation. ARI quantifies the alignment between the predicted and ground-truth segmentation masks, allowing us to compare inferred segmentation masks to ground-truth masks while ignoring background pixels. mIoU is a widely used segmentation metric that calculates the mean Intersection over Union values for different classes or objects in a segmentation task. In addition, to assess video quality, we report PSNR, SSIM \cite{wang2004image}, and LPIPS \cite{zhang2018unreasonable}. Additionally, for evaluating the model's prediction performance on fluid dynamics, we employ metrics such as Root Mean Square Error (RMSE), Mean Absolute Error (MAE), and High Correction Time (HCT).

\begin{figure*}[htbp]
\centering
\begin{minipage}{0.4\textwidth}
\begin{table}[H]
\caption{Evaluations of visual quality on OBJ3D dataset. * indicates our re-implementation.}
\label{tab4-1}
\vspace{-6pt}
\begin{tabular}{lccc}
\toprule
{\bf Model}&\small {\bf PSNR}&\small {\bf SSIM}&\small {\bf LPIPS$_{\times 100}$}$\downarrow$\\
\midrule
SAVi-Dyn&\textbf{32.94}&\textbf{0.91}&12.00\\
SlotFormer&32.41&\textbf{0.91}&8.00\\
STATM*&32.65&\textbf{0.91}&7.70\\
SlotPi (Ours)&32.67&\textbf{0.91}&\textbf{7.66}\\
\bottomrule
\end{tabular}
\vspace{-6pt}
\end{table}
\end{minipage}%
\hfill
\begin{minipage}{0.58\textwidth}
\begin{table}[H]
\caption{Evaluations of visual quality and object dynamics on CLEVRER. SlotFormer*
indicates the re-implemented results, where STATM-SAVi is used for slot extraction followed by training SlotFormer.}
\label{tab4-2}
\vspace{-6pt}
\begin{tabular}{lcccccc}
\toprule
{\bf Model}&\small {\bf PSNR}&\small {\bf SSIM}&\small {\bf LPIPS$_{\times 100}$}$\downarrow$&\small {\bf FG-ARI}\textsf{(\%)}&\small {\bf FG-mIoU}\textsf{(\%)}&\small {\bf ARI}\textsf{(\%)}\\
\midrule
SlotFormer*&30.21&0.89&11.20&63.10&49.27&63.25\\
STATM&30.22&0.89&10.78&64.56&49.57&63.30\\
SlotPi (Ours)&\textbf{30.41}&\textbf{0.90}&\textbf{10.51}&\textbf{66.12}&\textbf{50.32}&\textbf{64.16}\\
\bottomrule
\end{tabular}
\end{table}
\end{minipage}
\end{figure*}

\begin{figure*}[htbp]
\begin{small}
    \centering    
    \begin{minipage}[t]{1.0\linewidth}
    \centering
        \begin{tabular}{@{\extracolsep{\fill}}c@{}c@{}@{\extracolsep{\fill}}}
        (a) OBJ3D \hspace{7.2cm} (b) CLEVRER 
        \end{tabular}
    \end{minipage}
    \begin{minipage}[t]{1.0\linewidth}
    \centering
    \rotatebox{90}{Step} \
        \begin{tabular}{@{\extracolsep{\fill}}c@{}c@{}@{\extracolsep{\fill}}}
        \vspace{0.22cm}
        \hspace{0.38cm} 0 \hspace{0.7cm} 6 \hspace{0.7cm} 12 \hspace{0.7cm} 18 \hspace{0.7cm} 24 \hspace{0.7cm} 30 \hspace{0.7cm} 36 \hspace{0.7cm} 42 \hspace{0.75cm} 0 \hspace{0.7cm} 6 \hspace{0.7cm} 12 \hspace{0.7cm} 18 \hspace{0.7cm} 24 \hspace{0.7cm} 30 \hspace{0.7cm} 36 \hspace{0.7cm} 42 \hspace{0.38cm}
        \end{tabular}
    \end{minipage}
    \begin{minipage}[t]{1.0\linewidth}
    \centering
    \rotatebox[origin=c]{90}{G.T.} \
        \begin{tabular}{@{\extracolsep{\fill}}c@{}c@{}@{\extracolsep{\fill}}}
            \includegraphics[width=0.48\linewidth]{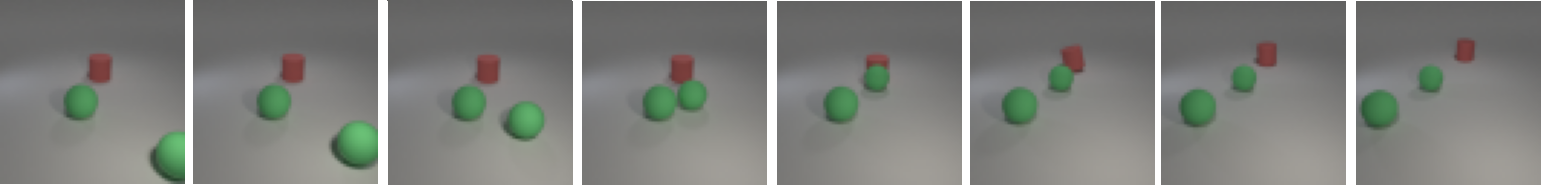}& \ 
            \hspace{0.05cm}
            \includegraphics[width=0.48\linewidth]{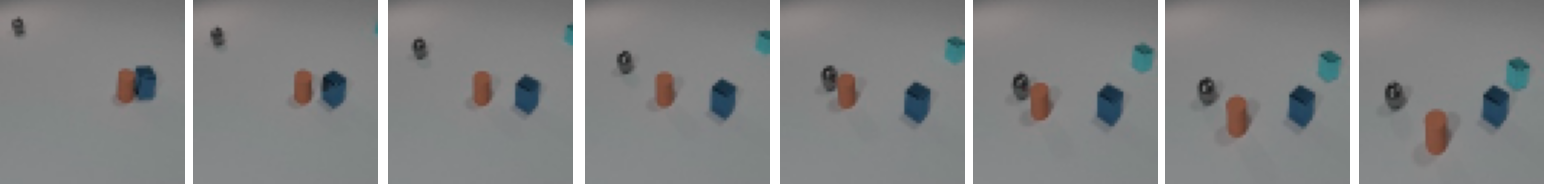}\\
        \end{tabular}
    \end{minipage}
    \begin{minipage}[t]{1.0\linewidth}
    \centering
    \rotatebox[origin=c]{90}{SlotFormer} \
        \begin{tabular}{@{\extracolsep{\fill}}c@{}c@{}@{\extracolsep{\fill}}}
            \includegraphics[width=0.48\linewidth]{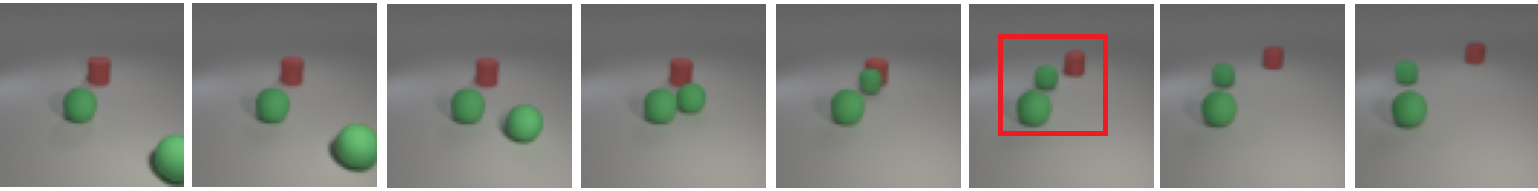}& \ 
            \hspace{0.05cm}
            \includegraphics[width=0.48\linewidth]{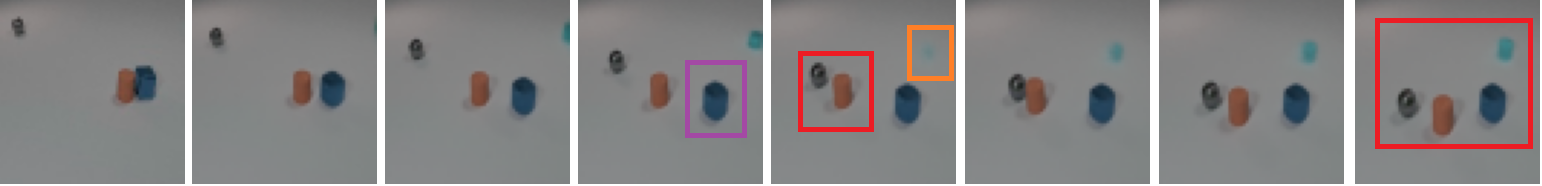}\\
        \end{tabular}
    \end{minipage}   
    \begin{minipage}[t]{1.0\linewidth}
    \rotatebox[origin=c]{90}{STATM} \
    \centering
        \begin{tabular}{@{\extracolsep{\fill}}c@{}c@{}@{\extracolsep{\fill}}}
            \includegraphics[width=0.48\linewidth]{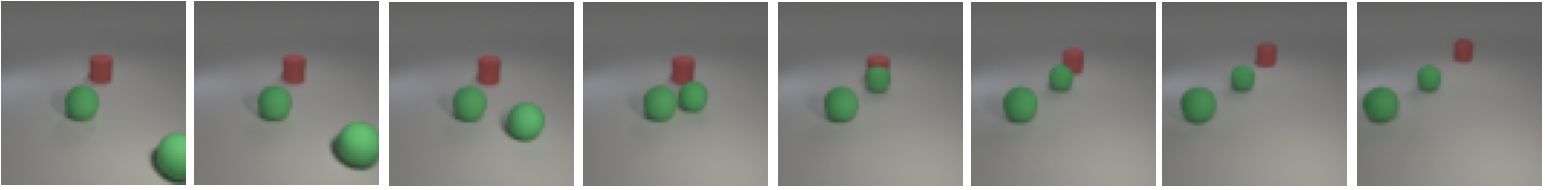}& \  
            \hspace{0.05cm}
            \includegraphics[width=0.48\linewidth]{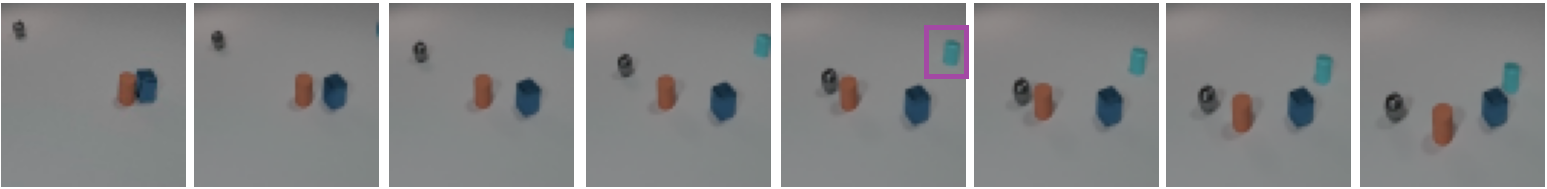}\\
        \end{tabular}
    \end{minipage}
    \begin{minipage}[t]{1.0\linewidth}
    \centering
    \rotatebox[origin=c]{90}{\textbf{Ours}} \
        \begin{tabular}{@{\extracolsep{\fill}}c@{}c@{}@{\extracolsep{\fill}}}
            \includegraphics[width=0.48\linewidth]{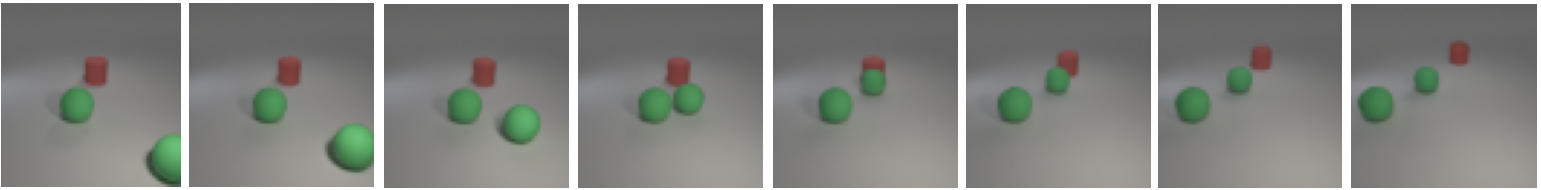}& \  
            \hspace{0.05cm}
            \includegraphics[width=0.48\linewidth]{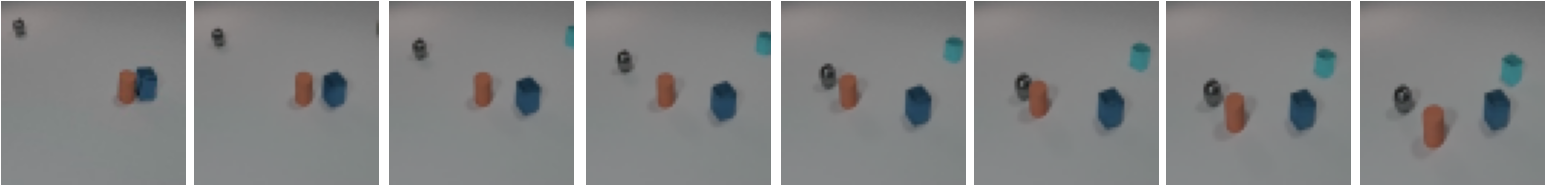}\\
        \end{tabular}
    \end{minipage}   
\vspace{-6pt}
\caption{Long-sequence prediction results on OBJ3D and CLEVRER. On OBJ3D, STATM performs well, but Slotformer exhibits incorrect dynamics (red boxes). On the more complex CLEVRER dataset, Slotformer's performance worsens, as its predictions begin to deviate from the ground truth, exhibiting artifacts such as blurriness (orange boxes), incorrect dynamics (red boxes), and shape changes (purple boxes). STATM's predictions also show issues (purple boxes) . Meanwhile, our model demonstrates strong performance.}
\label{fig4-1}
\vspace{-6pt}
\end{small}
\end{figure*}

\textbf{Datasets.} We validate the predictive capability of our model on four datasets, CLEVRER, OBJ3D, the fluid dataset, and a real-world dataset, and demonstrates its ability for downstream VQA task on two datasets, CLEVRER and Physion.

\textit{CLEVRER} dataset \cite{yi2019clevrer} is specifically designed for video understanding and reasoning, featuring complex object interactions and the emergence of new objects. For the VQA task, CLEVRER includes four types of questions: descriptive ("what color"), explanatory ("what’s responsible for"), predictive ("what will happen next"), and counterfactual ("what if"). Predictive questions require the model to simulate future interactions between objects, such as collisions. Therefore, our primary focus is on improving the accuracy of predictive question responses.

\textit{OBJ3D} dataset \cite{lin2020improving} is similar to the CLEVRER dataset, but unlike CLEVRER, which initially contains multiple moving objects and smaller objects, OBJ3D is simpler. It starts with a moving ball and 3 to 5 static objects, with the ball moving toward the static objects to create interactions.

\textit{Physion} dataset \cite{bear2021physion} is a VQA dataset and benchmark featuring realistic simulations of eight physical phenomena, designed to evaluate models based on human intuition about how objects move and interact with one another. The goal of this dataset is to predict whether a red agent will make contact with a yellow patient as the scene progresses.

\textit{Fluid Dataset.} We use the Navier-Stokes (NS) equations as the foundation for our fluid dataset. Predicting the physical systems described by the NS equations is a significant challenge in fluid dynamics due to the complex spatiotemporal patterns involved. We utilize the dataset provided by PDE-Refiner \cite{lippe2023pde}, with a time step of $\Delta {t} = 7.0125\times10^{-3}$ and resolution of $2048\times2048$. To accommodate training needs, we downsample each frame's grid to $64\times64$ and time step to $16 \Delta t$. 

\textit{Real-world} dataset is create by us and consists of fluid and objects floating on its surface. The scenes in the dataset include indoor fixed containers, outdoor lakes, and other environments. We collected data at different locations and times to ensure that the dataset includes various conditions and fluid types. In indoor environments, we add different primary color dyes to the water to make the fluid appear differently. The objects in the dataset come in various shapes (such as cylinders, cubes, etc.), sizes, colors (such as red, yellow, blue, purple, etc.), and materials (such as wood, plastic, foam, etc.), with each scene containing 3 to 5 objects. At the beginning of each video, we apply an external force to some of the objects in the scene to give them different initial velocities. Unlike CLEVRER, the first frame of this dataset contains all the objects, without new objects appearing (though some may disappear). This approach is intended to help the model focus better on learning the motion patterns of the scene. The dataset includes 603 training videos and 25 validation videos, each sampled at 30 fps for over 10 seconds, with a resolution of $1920\times1080$. To facilitate training with object-centric models, we standardize the dataset by cropping all videos to 256 frames and resizing each frame to $192\times108$.

\textbf{Implementation and Setting.} To validate the robustness of our model, we use multiple object-centric models to extract slot features from the datasets for training SlotPi. We employ STATM-SAVi on the CLEVRER and real-world dataset. On the CLEVRER dataset, STATM-SAVi uses a CNN as the encoder, with the number of slots set to 7 and the dimension set to 128. On the real-world dataset, we replace the CNN encoder with ResNet \cite{he2016deep}, setting the number of slots to 6 and the dimension to 192 to extract more accurate slot features. We also use the pre-trained weights provided by Wu et al. \cite{wu2023SlotFormer}, including SAVi \cite{kipf2021conditional} for OBJ3D and STEVE \cite{singh2022simple} for Physion. 

For the fluid dataset, since no objects are present, we use a convolution-based image patch embedding encoder with a patch size of $2\times2$ to encode the frame into 512 features. This results in 1024 tokens, each with a 512-dimension for each frame. Consequently, we modify the slot dimensions to 512 and set the number of slots to 1024 in SlotPi to accommodate the encoded tokens.

Regarding SlotPi setting, we set the initial learning rate to $2\times 10^{-4}$ ($1\times 10^{-4}$ for the fluid experiments) and use Adam (AdamW for the fluid) to train SlotPi on two H800 GPUs (one A100 for the fluid). For more implementation and setting, please refer to Appendix Section~\ref{append-baseline} and ~\ref{append-setting}. 

\begin{figure*}[t!]
\begin{small}
\centering    
    \begin{minipage}[t]{1.0\linewidth}
    \centering
    \rotatebox{0}{Rollout step} \
        \begin{tabular}{@{\extracolsep{\fill}}c@{}c@{}@{\extracolsep{\fill}}}
        \hspace{0.7cm} 0 \hspace{0.96cm} 6 \hspace{0.9cm} 12 \hspace{0.9cm} 18 \hspace{0.9cm} 24 \hspace{0.9cm} 30 \hspace{0.9cm} 36 \hspace{0.9cm} 42 \hspace{0.9cm} 48 \hspace{0.9cm} 54 \hspace{0.9cm} 60 \hspace{0.4cm}
        \end{tabular}
    \end{minipage}
    \begin{minipage}[t]{1.0\linewidth}
    \centering
    \rotatebox[origin=c]{0}{\hspace{0.5cm} G.T. (u) \hspace{0.1cm}} \
        \begin{tabular}{@{\extracolsep{\fill}}c@{}@{\extracolsep{\fill}}}
        \includegraphics[width=0.8\linewidth]{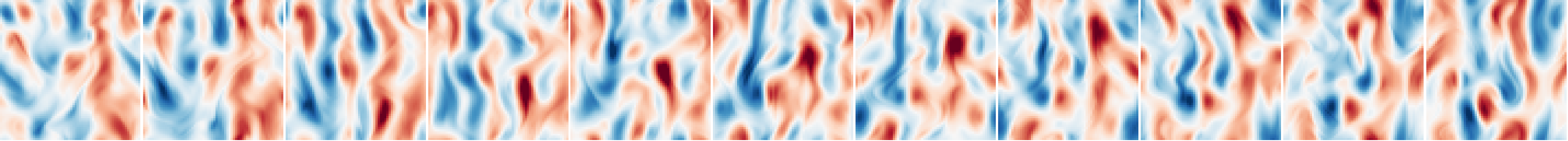}\\
        \end{tabular}
    \end{minipage}
    \begin{minipage}[t]{1.0\linewidth}
    \centering
    \rotatebox[origin=c]{0}{\hspace{0.5cm} Ours(u) \hspace{0.1cm}} \
        \begin{tabular}{@{\extracolsep{\fill}}c@{}@{\extracolsep{\fill}}}
        \includegraphics[width=0.8\linewidth]{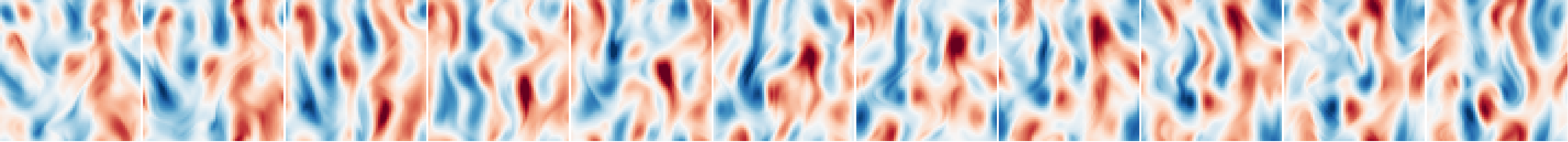}\\
        \end{tabular}
    \end{minipage}   
    \vspace{1pt} \\
    \begin{minipage}[t]{1.0\linewidth}
    \rotatebox[origin=c]{0}{\hspace{0.5cm} G.T. (v) \hspace{0.1cm}} \
    \centering
        \begin{tabular}{@{\extracolsep{\fill}}c@{}@{\extracolsep{\fill}}}
        \includegraphics[width=0.8\linewidth]{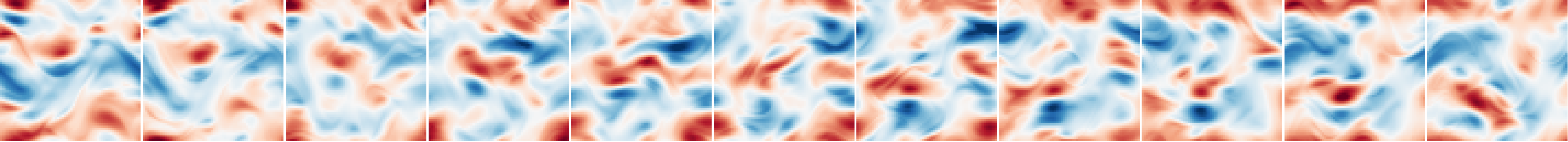}\\
        \end{tabular}
    \end{minipage}
    \begin{minipage}[t]{1.0\linewidth}
    \centering
    \rotatebox[origin=c]{0}{\hspace{0.5cm} Ours(v) \hspace{0.1cm}} \
        \begin{tabular}{@{\extracolsep{\fill}}c@{}@{\extracolsep{\fill}}}
        \includegraphics[width=0.8\linewidth]{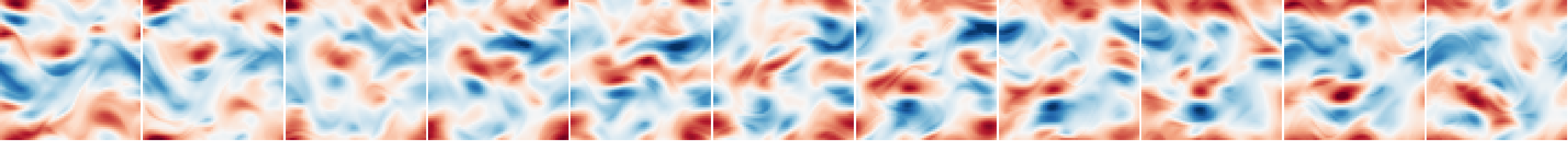}\\
        \end{tabular}
    \end{minipage}  
\caption{Rollout qualitative results on the NS fluid dataset.}
\label{fig4-4}
\vspace{-0pt}
\end{small}
\end{figure*}

\subsection{Benchmark Dataset Experiments}
\label{exp-41}

Table~\ref{tab4-1} and ~\ref{tab4-2} present the evaluation results for visual quality on OBJ3D and CLEVRER. As shown in Table~\ref{tab4-1}, on the OBJ3D dataset, our model achieves the best results on LPIPS compared to other models, and also performs well on PSNR and SSIM. In the more challenging CLEVRER dataset, where the objects are smaller and the interactions are more complex, our model outperforms SlotFormer by 0.2 in PSNR, 0.1 in SSIM, and 0.89 in LPIPS$\times$100, with these advantages being more pronounced than on the OBJ3D dataset. In the last three rows of Table~\ref{tab4-2}, we present the evaluation results for object dynamics on CLEVRER. We can see that our model outperforms all other models across all metrics, especially in FG-ARI, where we exceed STATM and Slotformer by 1.6 and 3.02 points. These results demonstrate that our model not only improves the quality of generated images but also provides more accurate dynamic simulations of objects. Furthermore, the advantages of our model become more evident as the complexity of the scene increases.

Figure~\ref{fig4-1} shows the generation results on OBJ3D and CLEVRER. As observed, on the simpler OBJ3D dataset, both STATM and our model perform well, while SlotFormer exhibits incorrect dynamics. On the more complex CLEVRER dataset, SlotFormer’s performance deteriorates further, exhibiting blurriness, incorrect dynamics, and inaccurate colors, and even becoming completely inconsistent with the ground truth. While STATM shows good dynamic simulation results, it still suffers from issues like object deformation. In contrast, our model continues to perform well. Please refer to Appendix Figure~\ref{fig-s2} and ~\ref{fig-s3} for more qualitative results.

\begin{table}[t!]
\caption{Predictive VQA accuracy on CLEVRER.}
\label{tab4-3}
\vspace{-10pt}
\begin{tabular}{lcc}
\toprule
{\bf Model}&{\bf per opt.}(\%)&{\bf per ques.}(\%)\\
\midrule
VRDP&95.68&91.35\\
SlotFormer&96.50&93.29 \\
STATM&96.62&93.63 \\
SlotPi (Ours)&\textbf{96.72}&\textbf{93.68} \\
\bottomrule
\end{tabular}
\vspace{-6pt}
\end{table}

In Table~\ref{tab4-3} and ~\ref{tab4-4}, we present the VQA results on the CLEVRER and Physion datasets. From the results in Table~\ref{tab4-3}, it is clear that, as an unsupervised object-centric prediction method, our model significantly outperforms the previous state-of-the-art STATM. Furthermore, on the publicly available CLEVRER leaderboard for the predictive question subset, our approach achieved first place in the per option setting and second rand in the per question setting. More detailed results can be found in the Appendix Table~\ref{tab-s1}. In Table~\ref{tab4-4}, we observe that on the more challenging Physion dataset, our model outperforms all other baselines, including scene-centric diffusion models like VDT. For qualitative results, please refer to Appendix Figure~\ref{fig-s1}. These results collectively demonstrate that our model's video predictions are more physically consistent.

\textbf{Discussion.} Clearly, our model demonstrates distinct advantages in prediction and downstream VQA tasks on the current inference-related benchmark datasets.

\begin{table}[t!]
\caption{Predictive VQA accuracy on Physion-Collide. * indicates our re-implementation}
\label{tab4-4}
\vspace{-6pt}
\begin{tabular}{lc}
\toprule
{\bf Model}&\small {\bf Accuracy}(\%)\\
\midrule
Human (upper bound)&80.0\\
VDT&65.3\\
SlotFormer&69.3\\
STATM*&72.1\\
SlotPi (Ours)&\textbf{74.4}\\
\bottomrule
\end{tabular}
\vspace{-6pt}
\end{table}

\subsection{Fluid Dataset Experiments}
\label{exp-42}

\begin{figure*}[t!]
\centering
\begin{minipage}{0.36\textwidth}
\begin{table}[H]
\caption{Results on the NS fluid dataset.}
\label{tab4-5}
\vspace{-6pt}
\begin{tabular}{lccc}
\toprule
{\bf Model}&\small {\bf RMSE} $\downarrow$&\small {\bf MAE} $\downarrow$&\small {\bf HCT (s)}\\
\midrule
FNO&0.8306&0.6362&2.6928\\
U-net&0.4393&0.3220&7.5174\\
SlotPi (Ours)&\textbf{0.2816}&\textbf{0.1887}&\textbf{8.976}\\
\bottomrule
\end{tabular}
\end{table}
\end{minipage}%
\hfill
\begin{minipage}{0.62\textwidth}
\begin{table}[H]
\caption{Metrics of visual quality and object dynamics on the real-world dataset.}
\label{tab4-6}
\vspace{-6pt}
\begin{tabular}{lcccccc}
\toprule
{\bf Model}&\small {\bf PSNR}&\small {\bf SSIM}&\small {\bf LPIPS} $\downarrow$&\small {\bf FG-ARI} \textsf{(\%)}&\small {\bf FG-mIoU} \textsf{(\%)}&\small {\bf ARI} \textsf{(\%)}\\
\midrule
SlotFormer&26.48&0.87&0.22&67.61&52.32&63.68\\
STATM&28.33&0.89&0.20&72.91&58.23&68.74\\
SlotPi (Ours)&\textbf{29.54}&\textbf{0.90}&\textbf{0.19}&\textbf{76.98}&\textbf{63.27}&\textbf{72.78}\\
\bottomrule
\end{tabular}
\end{table}
\end{minipage}
\end{figure*}

\begin{figure*}[t!]
\begin{small}
\centering    
    \begin{minipage}[t]{1.0\linewidth}
    \centering
    \rotatebox{0}{Rollout step} \
        \begin{tabular}{@{\extracolsep{\fill}}c@{}c@{}@{\extracolsep{\fill}}}
        \hspace{0.7cm} 0 \hspace{1.1cm} 6 \hspace{1.05cm} 12 \hspace{1.05cm} 18 \hspace{1.05cm} 24 \hspace{1.05cm} 30 \hspace{1.05cm} 36 \hspace{1.05cm} 42 \hspace{1.05cm} 48 \hspace{1.05cm} 54 \hspace{0.6cm}
        \end{tabular}
    \end{minipage}
    \begin{minipage}[t]{1.0\linewidth}
    \centering
    \rotatebox[origin=c]{0}{\hspace{0.5cm} G.T. \hspace{0.1cm}} \
        \begin{tabular}{@{\extracolsep{\fill}}c@{}@{\extracolsep{\fill}}}
        \includegraphics[width=0.802\linewidth]{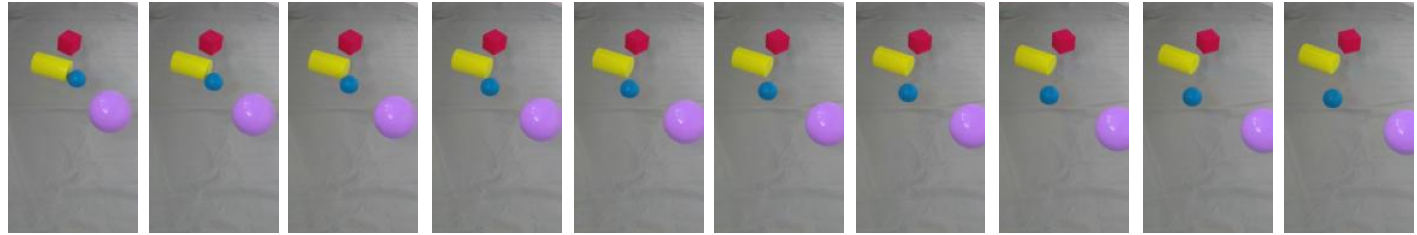}\\
        \end{tabular}
    \end{minipage}
    \begin{minipage}[t]{1.0\linewidth}
    \centering
    \rotatebox[origin=c]{0}{\hspace{0.5cm} Ours \hspace{0.1cm}} \
        \begin{tabular}{@{\extracolsep{\fill}}c@{}@{\extracolsep{\fill}}}
        \includegraphics[width=0.8\linewidth]{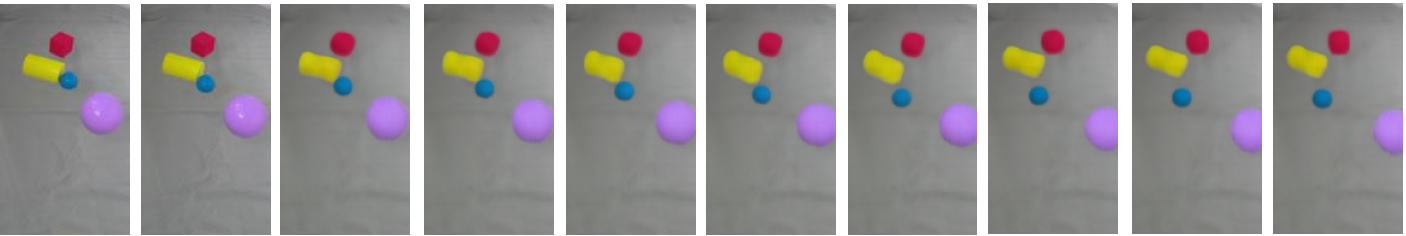}\\
        \end{tabular}
    \end{minipage}   
\caption{Prediction results on real-world dataset. The fluid flows slowly downward, influencing each object to move slowly in the same direction. Initially, we impart an upward velocity to the yellow object, causing it to move upward. However, the presence of the red object, which impacts the fluid dynamics and subsequently the yellow object, not only prevents it from moving entirely upward but also causes it to rotate. Meanwhile, the presence of the yellow object affects the red object, causing it not to drift with the fluid but to remain almost stationary. Our model has successfully predicted these complex scene dynamics.}
\label{fig4-6}
\end{small}
\end{figure*}

On the fluid dataset, we employ a convolution-based image patch embedding encoder with a patch size of $2\times2$ to encode each frame into 512 features. Additionally, we adjust the slot dimensions to 512 and set the number of slots to 1024 in the SlotPi model.

We trained our model using 30 trajectories of NS with diverse initial conditions and tested it using 10 trajectories. The quantitative results are displayed in Table~\ref{tab4-5}. These results demonstrate that our model significantly outperforms other models on RMSE, MAE, and HCT metrics. Furthermore, the qualitative analysis shown in Figure~\ref{fig4-4} indicates that our model's predictions closely align with the ground truth values. These results confirm that our model is highly effective at predicting fluid dynamics.

\textbf{Discussion.} Clearly, our model is capable of performing well on the NS fluid dataset, which indicates that it is effective not only in scenarios involving object interactions but also excels in the fluid dynamics simulations without objects. This demonstrates the model's excellent adaptability and generalization capabilities.

\subsection{Real-world Dataset Experiments} 
\label{exp-43}

\begin{figure}[t!]
\begin{center}
\centerline{\includegraphics[width=0.95\linewidth]{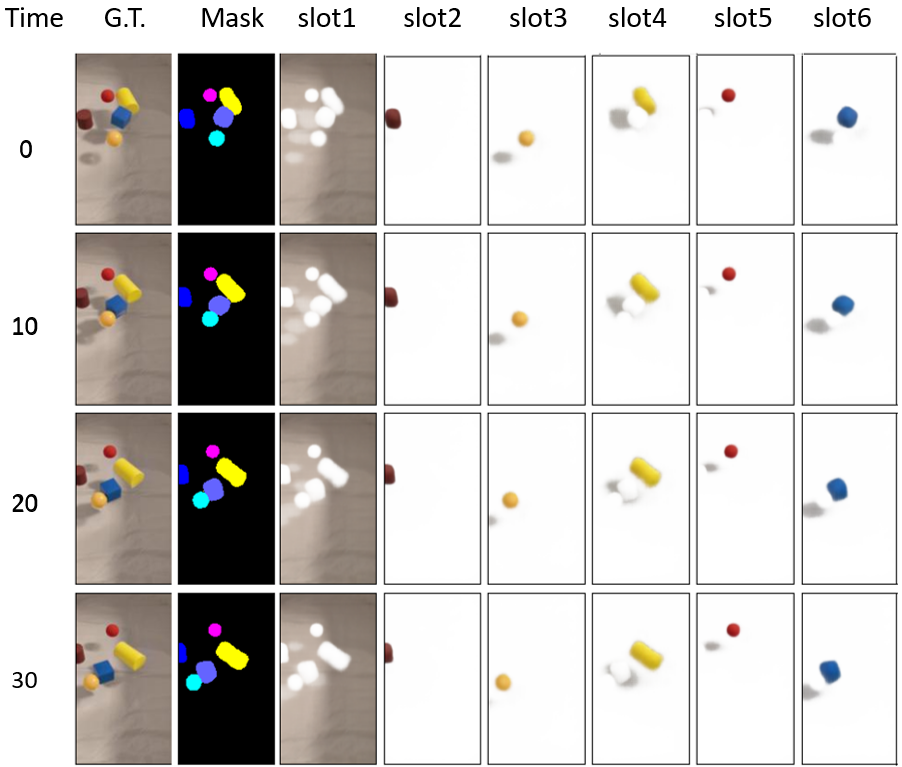}}
\vspace{-6pt}
\caption{Quality results of scene decomposition on real-world dataset over time. Mask is predicted by STATM-SAVi. The features of each object are decomposed into separate slot, and these slots do not become disordered over time.} 
\label{fig4-5}
\vspace{-10pt}
\end{center}
\end{figure}

To better decompose the scenes in the real-world dataset, we replaced the encoder of STATM-SAVi with ResNet \cite{he2016deep} and adjusted the slot dimensions to 192 for training on the real-world dataset. Figure~\ref{fig4-5} shows the quality results of scene decomposition. We observe that the model successfully decomposes the features related to each object into a separate slot, and over time, each slot remains consistent without confusion. For more decomposition results, please refer to Appendix Table~\ref{tab-s2} and Figure~\ref{fig-s4}.

Table~\ref{tab4-6} presents the prediction results on the dataset. We find that our model significantly outperforms others in terms of both image quality and dynamic simulation. From the generation results in Figure~\ref{fig4-6}, it is evident that our model's simulated result closely align with the ground truth, even capturing behaviors such as the rotation of objects on the water surface. These results demonstrate that our model can effectively simulate the dynamics of the scene.

\textbf{Discussion.} Certainly, our model performs well on the real-world dataset, which include object interactions (Section~\ref{exp-41}) and fluid dynamics (Section~\ref{exp-42}), as well as interactions between fluids and objects. This performance on the dataset further demonstrates our model's robust predictive capabilities and adaptability.

\subsection{Ablation Study}
\label{exp-44}

\begin{figure}[t!]
\begin{center}
\centerline{\includegraphics[width=0.98\linewidth]{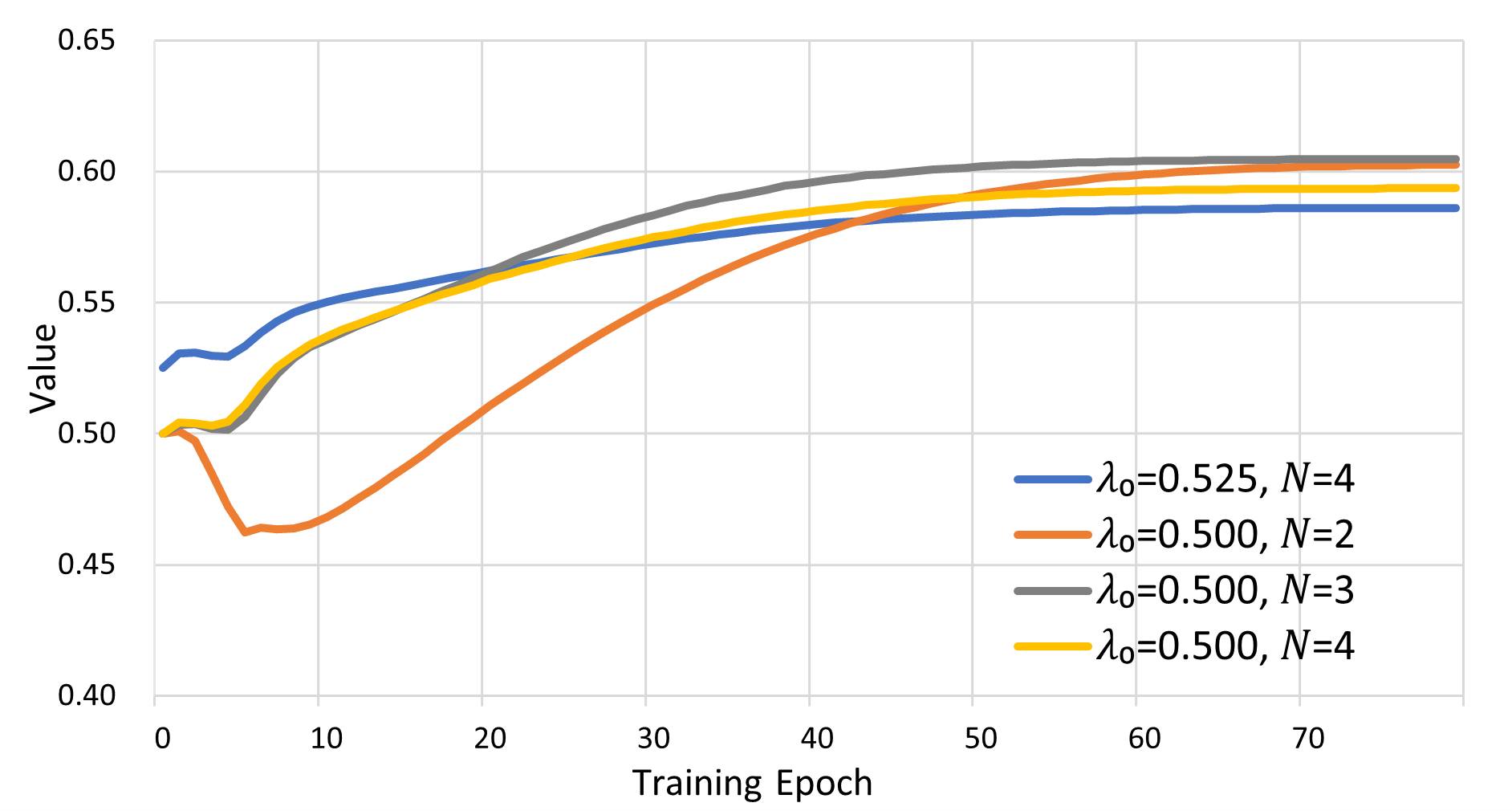}}
\vspace{-6pt}
\caption{The value of the learnable $\lambda$ changes throughout the training process. $\lambda_{0}$ represents the initial value of $\lambda$; N denotes the number of attention blocks in the spatiotemporal reasoning module.} 
\label{fig4-7}
\vspace{-6pt}
\end{center}
\end{figure}

\textbf{Physical Module.} We modified the value of $\lambda$ in Equation~(\ref{eq6}) to investigate the impact of the physical module on the overall model performance. To this end, we designed two experimental setups: 1) We set $\lambda$ as a learnable parameter between 0 and 1 and trained the model to observe the variations in the $\lambda$ value, and 2) We fixed $\lambda$ at 0.1 and 1 to observe the model's performance under these conditions. As shown in Table~\ref{tab4-7}, the model's performance is slightly better when $\lambda$=1. From Figure~\ref{fig4-7}, it can be observed that as the training epochs progress, the value of $\lambda$ gradually increases, but it does not exceed 1. When the physical module is removed (i.e., $\lambda$=0), the model degrades to a structure similar to STATM or Slotformer, resulting in inferior performance. Therefore, we set $\lambda$=1 by default for our experiments.

\begin{table}[t!]
\caption{Evaluations of object dynamics on CLEVRER with different values of $\lambda$.}
\label{tab4-7}
\vspace{-6pt}
\begin{tabular}{lccc}
\toprule
{\bf Values}&\small {\bf FG-ARI}\textsf{(\%)}&\small {\bf FG-mIoU}\textsf{(\%)}&\small {\bf ARI}\textsf{(\%)}\\
\midrule
$\lambda$=1&\textbf{66.12}&\textbf{50.32}&\textbf{64.16}\\
$\lambda$=0.1&65.18&49.91&63.86\\
learnable&65.55&50.10&63.89\\
\bottomrule
\end{tabular}
\vspace{-6pt}
\end{table}

\begin{table}[t!]
\caption{Evaluations of different embedding methods on CLEVRER. \textquoteleft MLP\textquoteright\ refers to approach of using an MLP to solve for the generalized momentum $P$, generalized coordinates $Q$, and scene energy, embedding Hamiltonian within the model. \textquoteleft Attention\textquoteright\  refers to use of attention mechanisms for task. }
\label{tab4-8}
\vspace{-6pt}
\begin{tabular}{lccc}
\toprule
{\bf Method}&\small {\bf FG-ARI}\textsf{(\%)}&\small {\bf FG-mIoU}\textsf{(\%)}&\small {\bf ARI}\textsf{(\%)}\\
\midrule
MLP&65.08&49.79&63.61\\
Attention&\textbf{66.12}&\textbf{50.32}&\textbf{64.16}\\
\bottomrule
\end{tabular}
\vspace{-6pt}
\end{table}

We experimented with different approaches to embed the Hamiltonian equation within the model: 1) embedding using a multilayer perceptron (MLP), and 2) embedding using an attention mechanism. The results are shown in Table~\ref{tab4-8}. We found that the attention mechanism outperformed the MLP approach for embedding the Hamiltonian equation. The MLP required multiple layers (7 or 9) to solve for the generalized momentum $P$, generalized coordinates $Q$, and scene energy, which significantly increased the model's parameter count. In contrast, the attention mechanism was able to solve for $P$, $Q$, and the scene energy with just one attention block. Therefore, we use the attention mechanism for embedding by default. For more detailed results, please refer to Appendix Table~\ref{tab-s4}.

\begin{figure}[t!]
\begin{small}
\centering    
    \begin{minipage}[t]{1.0\linewidth}
    \centering
    \rotatebox[origin=c]{90}{G.T.} \
        \begin{tabular}{@{\extracolsep{\fill}}c@{}@{\extracolsep{\fill}}}
        \includegraphics[width=0.9\linewidth]{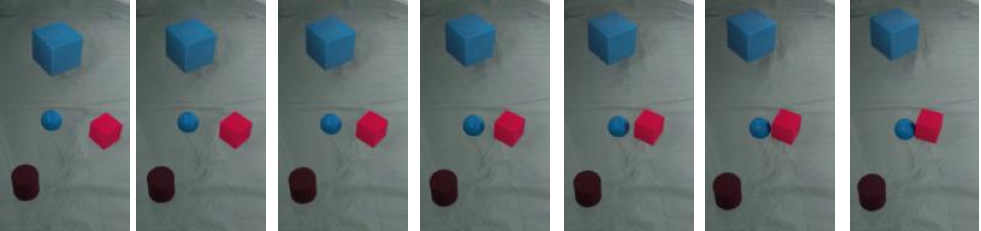}\\
        \end{tabular}
    \end{minipage}
    \begin{minipage}[t]{1.0\linewidth}
    \centering
    \rotatebox[origin=c]{90}{Ours (10to15)} \
        \begin{tabular}{@{\extracolsep{\fill}}c@{}@{\extracolsep{\fill}}}
        \includegraphics[width=0.9\linewidth]{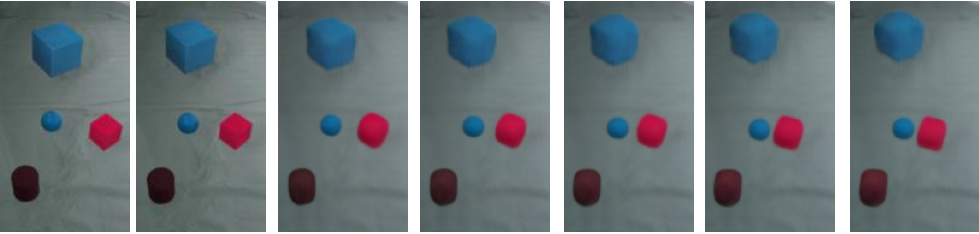}\\
        \end{tabular}
    \end{minipage}   
    \begin{minipage}[t]{1.0\linewidth}
    \centering
    \rotatebox[origin=c]{90}{Ours (6to10)} \
        \begin{tabular}{@{\extracolsep{\fill}}c@{}@{\extracolsep{\fill}}}
        \includegraphics[width=0.9\linewidth]{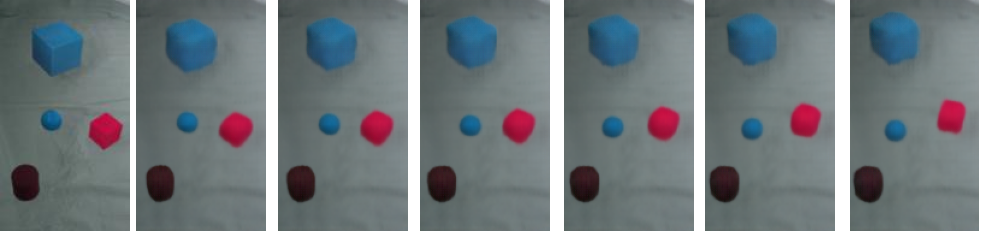}\\
        \end{tabular}
    \end{minipage}  
\caption{Prediction results of different number of frames for training. \textquoteleft6 to 10\textquoteright\ refers to using 6 burn-in and 10 roll-out frames for training the model. With this setup, the model struggles to capture the motion of objects slowly drifting with the fluid, mistakenly predicting that the red object and the blue ball are almost stationary. In contrast, using 10 burn-in and 15 roll-out frames results in better performance. }
\label{fig4-8}
\vspace{-8pt}
\end{small}
\end{figure}

\textbf{Number of frames for training}. On our real-world dataset, we trained the model using different numbers of input frames. The experimental results show that the combination of 6 burn-in frames and 10 roll-out frames performs worse than the combination of 10 burn-in frames and 15 roll-out frames. By examining the generated results Figure~\ref{fig4-8}, we found that the movement of objects floating in the scene may be very slow due to the influence of the fluid. In this case, with fewer input frames, the model struggles to effectively learn the motion of the objects, mistakenly predicting them as almost stationary. Therefore, on the real-world dataset, we use 10 burn-in frames and 15 roll-out frames for training the model.

\subsection{Limitations}

The model needs to be retrained on each dataset in order to function effectively. Currently, our model primarily follows a two-stage training process, where slots need to be extracted in advance using object-centric methods. Additionally, the model's configuration may vary across datasets. We would prefer to have a unified architecture that can be jointly trained on more datasets, combining the base object-centric model and SlotPi, enabling it to perform effectively across different scenarios.

\section{Conclusion}

Humans can acquire knowledge by observing the world and use this knowledge to better understand it, this is an important aspect of intuitive physics. Intuitive physics enhances human adaptability, enabling quick judgments and reasoning about the dynamics of various scenarios around them. Our goal is to mimic human intuitive physical behavior by constructing an object-centric prediction model that performs well in diverse environments. We have designed a physical module that embeds universal physical laws of the physical world, ensuring that the model's predictions are constrained by fundamental physical laws. Additionally, we introduced a spatiotemporal reasoning module to capture and reason about parts of the system that cannot be directly inferred from the physical module. This not only addresses the limitation of traditional physical models that only handle conservative systems but also provides physical constraints for the overall model's predictions, ensuring they adhere to basic physical laws. Through a series of experiments, we demonstrated the advantages of our model in tasks such as prediction and Visual Question Answering (VQA) on benchmark datasets. Experiments on fluids show that our model excels in making predictions about fluids. Furthermore, we created a real-world dataset that includes object interactions and fluid dynamics, as well as interactions between fluids and objects, and validated our model's capabilities on this dataset. All these indicate the excellent adaptability of our model, capable of accommodating a variety of predictive scenarios. Although there are still many challenges in this field, our work provides a research foundation for building more adaptable world models.

\begin{acks}
The work is supported by the National Natural Science Foundation of China (No. 92270118, No. 62276269) and the Beijing Natural Science Foundation (No. 1232009).
\end{acks}

\bibliographystyle{ACM-Reference-Format}
\balance
\bibliography{sample-base}


\clearpage
\appendix
\balance

\setcounter{figure}{0}
\setcounter{table}{0}
\renewcommand{\thefigure}{S\arabic{figure}}
\renewcommand{\thetable}{S\arabic{table}}

\begin{center}
    {\Large \textbf{Appendix}}
\end{center}

This supplementary material file provides the appendix section to the main article.

\section{Baselines}
\label{append-baseline}

To validate the effectiveness of our model, we selected appropriate baseline models for comparison based on the datasets.

\textbf{SlotFormer.} SlotFormer \cite{wu2023SlotFormer} is a transformer-based framework specifically designed for object-centric visual simulation. It utilizes slots extracted by pretrained object-centric models for making predictions. For the CLEVRER predictive VQA questions, SlotFormer is deployed, and Aloe is executed on the predicted slots. For the Physion VQA task, it employs a linear readout model trained on the predicted slots to evaluate the success of tasks.

On the OBJ3D dataset, we maintain the same settings as the official configurations. For the CLEVRER dataset, we use STATM-SAVi to extract slots to train SlotFormer, keeping all other settings consistent with the official implementation.

On the real-world dataset, we employ four transformer encoders for prediction, set the number of slots to 6, and slot dimensions to 192. We subsample the video by a factor of 2 and train for 60 epochs using the Adam optimizer, with a batch size of 64 and a learning rate of $2\times 10^{-4}$. Other training strategies are almost consistent with those used on other datasets.

\textbf{STATM.} STATM \cite{li2024reasoningenhanced} comprises two main components: a memory buffer and a slot-based time-space transformer. The memory buffer is responsible for storing slot information from upstream modules, while the time-space transformer leverages this information for temporal and spatial inference. There are several methods to integrate the results of temporal and spatial reasoning, and we utilize the default summation method as described in the paper. 

On the CLEVRER dataset, we subsample the video by a factor of 2 for training STATM, conducting approximately 500,000 training steps. On the OBJ3D dataset, we train for approximately 160,000 steps. For the real-world dataset, we also subsample the video by a factor of 2 and train for 60 epochs. Training across all datasets is conducted using 2 H800 GPUs, and we set the size of the memory buffer to match the number of burn-in frames. We utilize the Adam optimizer with a batch size of 64 and a learning rate of $2\times 10^{-4}$, applying the same warm-up and decay learning rate schedule for the first 2.5\% of the total training steps. Other training strategies and settings are consistent with those used for SlotFormer.

\textbf{Savi-dyn.} In SlotFormer \cite{wu2023SlotFormer}, Wu et al. use Transformer-LSTM module in PARTS \cite{zoran2021parts} replacing the Transformer predictor in SAVi.

\textbf{VRDP.} VRDP \cite{ding2021dynamic} effectively utilizes videos and language to learn visual concepts and infer physics models of objects and their interactions. It comprises three primary modules: a visual perception module, a concept learner, and a differentiable physics engine. The visual perception module is designed to parse scenes into object-centric representations. The concept learner then extracts prior knowledge from these object-centric representations based on the language. The physics engine uses this prior knowledge to perform dynamic simulations. We have implemented VRDP with a visual perception module that is trained based on object properties.

\textbf{VDT.} VDT \cite{lu2023vdt} is a transformer-based diffusion model for videos. The model features transformer blocks with modularized temporal and spatial attention modules, allowing it to efficiently capture long-range dependencies and temporal dynamics. VDT employs a novel approach where different segments or elements within a video frame are masked and predicted. We maintain settings consistent with the official configurations.

\textbf{FNO.} The Fourier Neural Operator (FNO) \cite{li2020fourier} is a deep learning model meticulously crafted to efficiently solve parametric partial differential equations (PDEs). FNO synergistically integrates neural networks with Fourier transforms and is structured into two main components. The first component primarily transforms input functions into the frequency domain, facilitating the extraction of global features. The second component comprises a local linear transformation aimed at extracting local features. This dual-component architecture enables FNO to capture both the broad patterns and intricate details necessary for accurately modeling and solving complex systems governed by PDEs. We utilize the Adam optimizer, with a learning rate of $1\times 10^{-3}$ and a batch size of 20. All other settings are consistent with those specified in the original paper. 

\textbf{UNet.} UNet \cite{ronneberger2015u} adopts an encoder-decoder structure where the encoder extracts features at multiple scales via a series of downsampling operations, while the decoder reconstructs the original image size through a series of upsampling steps. Skip connections are used to merge feature maps from corresponding layers, thus preserving local detail and capturing global context effectively. We utilize the modern UNet \cite{gupta2022towards} architecture with default settings.

\section{Implementation and Setting}
\label{append-setting}

\begin{figure*}[t!]
\begin{small}
    \centering    
    \begin{minipage}[t]{1.0\linewidth}
    \centering
    \rotatebox[origin=c]{90}{G.T.} \
        \begin{tabular}{@{\extracolsep{\fill}}c@{}@{\extracolsep{\fill}}}
        \includegraphics[width=0.96\linewidth]{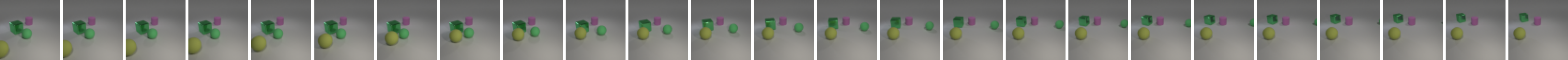}\\
        \end{tabular}
    \end{minipage}
    \begin{minipage}[t]{1.0\linewidth}
    \centering
    \rotatebox[origin=c]{90}{Ours} \
        \begin{tabular}{@{\extracolsep{\fill}}c@{}@{\extracolsep{\fill}}}
        \includegraphics[width=0.96\linewidth]{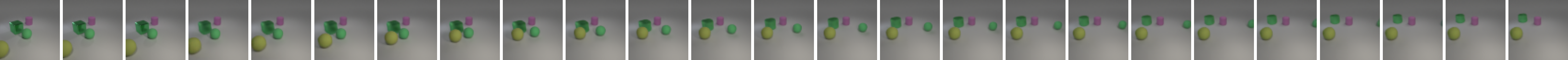}\\
        \end{tabular}
    \end{minipage}   
    \vspace{1pt} \\
    \begin{minipage}[t]{1.0\linewidth}
    \rotatebox[origin=c]{90}{G.T.} \
    \centering
        \begin{tabular}{@{\extracolsep{\fill}}c@{}@{\extracolsep{\fill}}}
        \includegraphics[width=0.96\linewidth]{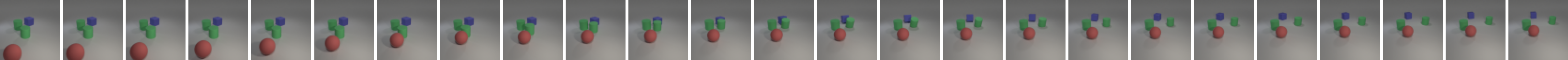}\\
        \end{tabular}
    \end{minipage}
    \begin{minipage}[t]{1.0\linewidth}
    \centering
    \rotatebox[origin=c]{90}{Ours} \
        \begin{tabular}{@{\extracolsep{\fill}}c@{}@{\extracolsep{\fill}}}
        \includegraphics[width=0.96\linewidth]{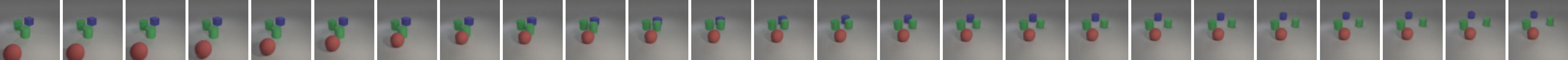}\\
        \end{tabular}
    \end{minipage}  
    \vspace{1pt} \\
    \begin{minipage}[t]{1.0\linewidth}
    \centering
    \rotatebox[origin=c]{90}{G.T.} \
        \begin{tabular}{@{\extracolsep{\fill}}c@{}@{\extracolsep{\fill}}}
        \includegraphics[width=0.96\linewidth]{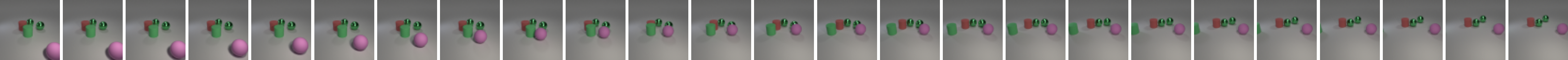}\\
        \end{tabular}
    \end{minipage}
    \begin{minipage}[t]{1.0\linewidth}
    \centering
    \rotatebox[origin=c]{90}{Ours} \
        \begin{tabular}{@{\extracolsep{\fill}}c@{}@{\extracolsep{\fill}}}
        \includegraphics[width=0.96\linewidth]{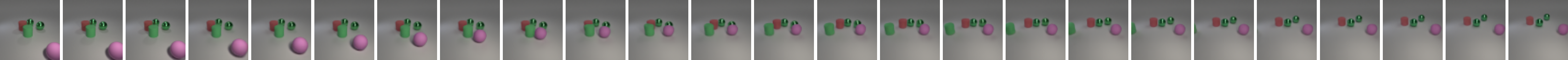}\\
        \end{tabular}
    \end{minipage}  
    \vspace{-6pt}
    \caption{Qualitative results of our model on the OBJ3D dataset.}
\label{fig-s2}
\vspace{-6pt}
\end{small}
\end{figure*}

\begin{figure*}[t!]
\begin{small}
    \centering    
    \begin{minipage}[t]{1.0\linewidth}
    \centering
    \rotatebox[origin=c]{90}{G.T.} \
        \begin{tabular}{@{\extracolsep{\fill}}c@{}@{\extracolsep{\fill}}}
        \includegraphics[width=0.96\linewidth]{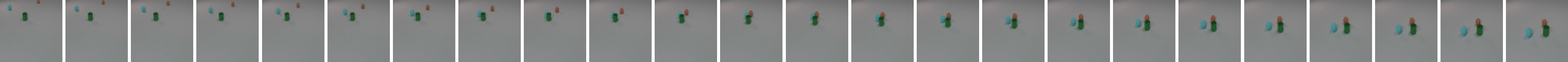}\\
        \end{tabular}
    \end{minipage}
    \begin{minipage}[t]{1.0\linewidth}
    \centering
    \rotatebox[origin=c]{90}{Ours} \
        \begin{tabular}{@{\extracolsep{\fill}}c@{}@{\extracolsep{\fill}}}
        \includegraphics[width=0.96\linewidth]{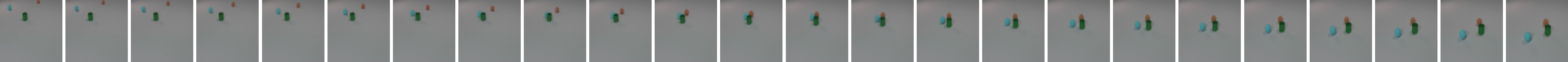}\\
        \end{tabular}
    \end{minipage}   
    \vspace{1pt} \\
    \begin{minipage}[t]{1.0\linewidth}
    \rotatebox[origin=c]{90}{G.T.} \
    \centering
        \begin{tabular}{@{\extracolsep{\fill}}c@{}@{\extracolsep{\fill}}}
        \includegraphics[width=0.96\linewidth]{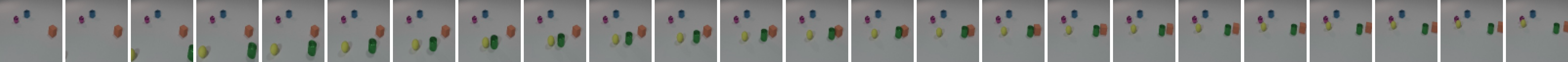}\\
        \end{tabular}
    \end{minipage}
    \begin{minipage}[t]{1.0\linewidth}
    \centering
    \rotatebox[origin=c]{90}{Ours} \
        \begin{tabular}{@{\extracolsep{\fill}}c@{}@{\extracolsep{\fill}}}
        \includegraphics[width=0.96\linewidth]{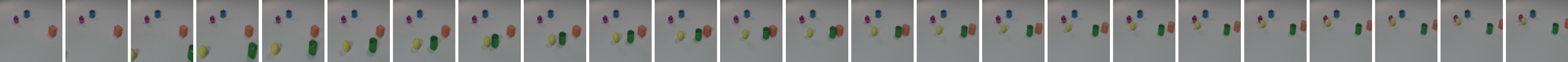}\\
        \end{tabular}
    \end{minipage}  
    \vspace{1pt} \\
    \begin{minipage}[t]{1.0\linewidth}
    \centering
    \rotatebox[origin=c]{90}{G.T.} \
        \begin{tabular}{@{\extracolsep{\fill}}c@{}@{\extracolsep{\fill}}}
        \includegraphics[width=0.96\linewidth]{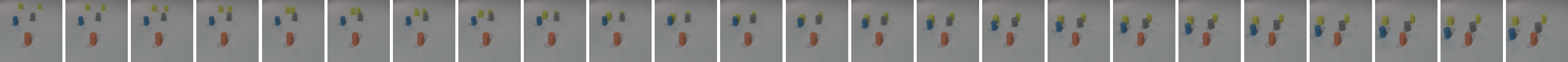}\\
        \end{tabular}
    \end{minipage}
    \begin{minipage}[t]{1.0\linewidth}
    \centering
    \rotatebox[origin=c]{90}{Ours} \
        \begin{tabular}{@{\extracolsep{\fill}}c@{}@{\extracolsep{\fill}}}
        \includegraphics[width=0.96\linewidth]{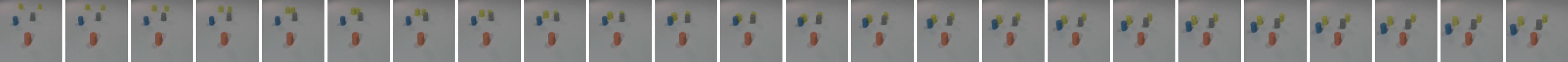}\\
        \end{tabular}
    \end{minipage}   
    \vspace{-6pt}
    \caption{Qualitative results of our model on the CLEVRER dataset.}
\label{fig-s3}
\vspace{-6pt}
\end{small}
\end{figure*}

\begin{table*}[ht!]
\renewcommand\arraystretch{1.1}
\caption{Accuracy on different questions and average results on CLEVRER.}
\vspace{-6pt}
\label{tab-s1}
\begin{center}
\begin{tabular}{lccclcclccc}
\toprule
\multicolumn{1}{l}{\bf \multirow{2}*{Model}} & \multicolumn{1}{c}{\bf \multirow{2}*{Descriptive}} & \multicolumn{2}{c}{\bf \multirow{1}*{Explanatory}} &\multicolumn{1}{l}{} & \multicolumn{2}{c}{\bf \multirow{1}*{Predictive}} &\multicolumn{1}{l}{} & \multicolumn{2}{c}{\bf \multirow{1}*{Counterfactual}} & \multicolumn{1}{l}{\bf \multirow{2}*{Average}} \\
\cline{3-4} \cline{6-7} \cline{9-10}  
{}&{}&per opt.&per ques.&{}&per opt.&per ques.&{}&per opt.&per ques.&{} \\
\hline 
VRDP&93.40&96.30&91.94&{}&95.68&91.35&{}&\textbf{94.83}&\textbf{84.29}&90.24 \\
SlotFormer&95.17&98.04&94.79&{}&96.50&93.29&{}&90.63&73.78&89.26 \\
STATM&\textbf{95.22}&98.15&95.04&{}&96.62&93.63&{}&90.57&73.90&89.44 \\
SlotPi (Ours)&95.16&\textbf{98.38}&\textbf{95.73}&{}&\textbf{96.72}&\textbf{93.68}&{}&91.07&75.13&89.93 \\
\bottomrule
\end{tabular}
\vspace{-6pt}
\end{center}
\end{table*}

We have reproduced STATM-SAVi using PyTorch to extract slots on CLEVRER dataset and real-world dataset. Structurally, we maintain consistency with the original STATM-SAVi, which includes an encoder, a decoder, a Slot Attention-based corrector, and a STATM predictor. On the CLEVRER dataset, we employ a simple CNN as the encoder. The slot size is set to 128, the number of slots is set to 7, and the spatial broadcast size of the decoder is $8\times8$. For the real-world dataset, we switch the encoder to ResNet-18, change the slot size to 192, set the number of slots to 6, and adjust the spatial broadcast size of the decoder to $12\times7$. We train STATM-SAVi for 21 epochs on the CLEVRER dataset and 46 epochs on the real-world dataset using 2$\times$A100 GPUs with 80GB memory each with a batch size of 64. On all datasets, we split each video into sub-sequences of 6 frames to train the model. We utilize the Adam optimizer with a learning rate of $2\times 10^{-4}$ and apply the same warm-up and decay learning rate schedule for the first 2.5\% of the total training steps.

For SlotPi, our training strategy on benchmark datasets aligns with those used for STATM and SlotFormer. For the fluid dataset, where no distinct objects are present, we contemplate utilizing either all slots or only the background slots to store the fluid features.So we just employ a convolution-based image patch embedding encoder with a $2\times2$ patch size to encode the frame into 512 features, resulting in 1024 tokens, each with a 512-dimension for each frame. Consequently, we adjust the slot dimensions in SlotPi to 512 and increase the number of slots to 1024 to accommodate these encoded tokens. We set the learning rate to $1\times 10^{-4}$ and use the AdamW optimizer. The scheduler employed is StepLR, with a step size of 100 and a gamma of 0.96. Other strategies remain consistent with those used for UNet.

On the real-world dataset, we employ four attention blocks within the spatiotemporal reasoning module for prediction. We configure the model with 6 slots, each having dimensions of 192. Training involves 10 burn-in frames and 15 rollout frames over 40 epochs, using Adam as the optimizer. The batch size is set to 64, with a learning rate of $2\times 10^{-4}$. We implement a warm-up and decay learning rate schedule for the initial 2.5\% of the total training steps to facilitate gradual learning improvements. This approach ensures that our training strategies are consistent with those utilized for SlotFormer and STATM, aiming to optimize performance and adaptability across diverse scenarios.

\section{Additional Experimental Results}
\label{append-result}

\textbf{Additional OBJ3D and CLEVRER Results.} We present additional qualitative results of longer time series on OBJ3D and CLEVRER in Figures~\ref{fig-s2} to~\ref{fig-s3}. It is evident that on the OBJ3D and CLEVRER datasets, our model not only accurately predicts the dynamics of objects but also maintains excellent image quality without issues like object disappearance or deformation.

\textbf{Additional VQA Results.} Detailed results about all questions on CLEVRER dataset are presented in Table~\ref{tab-s1}. We can see that our model not only performs optimally on predictive questions but also excels in other types of questions. 

\begin{figure*}[ht!]
\begin{small}
\centering    
    \begin{minipage}[t]{1.0\linewidth}
    \centering
    \rotatebox[origin=c]{0}{G. T.} \
        \begin{tabular}{@{\extracolsep{\fill}}c@{}@{\extracolsep{\fill}}}
        \includegraphics[width=0.9\linewidth]{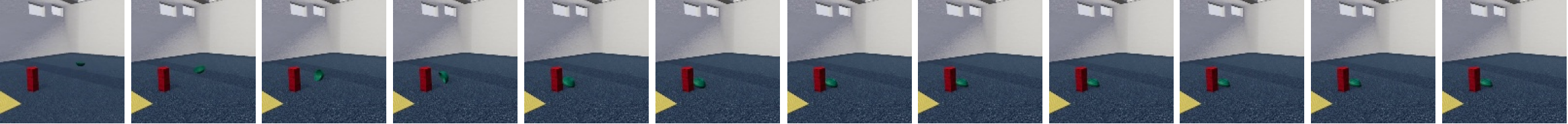}\\
        \end{tabular}
    \end{minipage}
    \begin{minipage}[t]{1.0\linewidth}
    \centering
    \rotatebox[origin=c]{0}{Ours} \
        \begin{tabular}{@{\extracolsep{\fill}}c@{}@{\extracolsep{\fill}}}
        \includegraphics[width=0.9\linewidth]{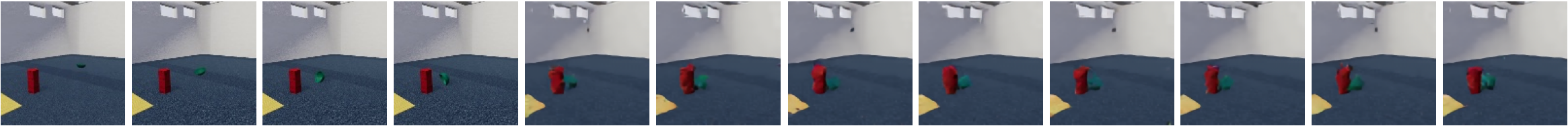}\\
        \end{tabular}
    \end{minipage}   
\caption{Qualitative results of Physion.}
\label{fig-s1}
\vspace{-6pt}
\end{small}
\end{figure*}

\textbf{Additional Physion Qualitative Results.} We show more qualitative results on Physion in Figure~\ref{fig-s1}. We observed that SlotPi can accurately simulate the motion of objects, but the image quality is low. This issue is \textit{due to the STEVE’s Transformer-based decoder, which is not designed for pixel-space reconstruction} \cite{wu2023SlotFormer}. Enhancing the decoding quality of the upstream model falls outside the research scope of this paper. 

\begin{table}[t!]
\caption{Quantitative results on real-world.}
\label{tab-s2}
\vspace{-6pt}
\begin{tabular}{lccc}
\toprule
{\bf Model}&\small {\bf FG-ARI}\textsf{(\%)}&\small {\bf FG-mIoU}\textsf{(\%)}&\small {\bf ARI}\textsf{(\%)}\\
\midrule
STATM-SAVi&96.60&84.57&90.43\\
\bottomrule
\end{tabular}
\vspace{6pt}
\end{table}

\begin{figure*}[t!]
\centering
    \begin{minipage}{0.48\textwidth}
        \includegraphics[width=\linewidth]{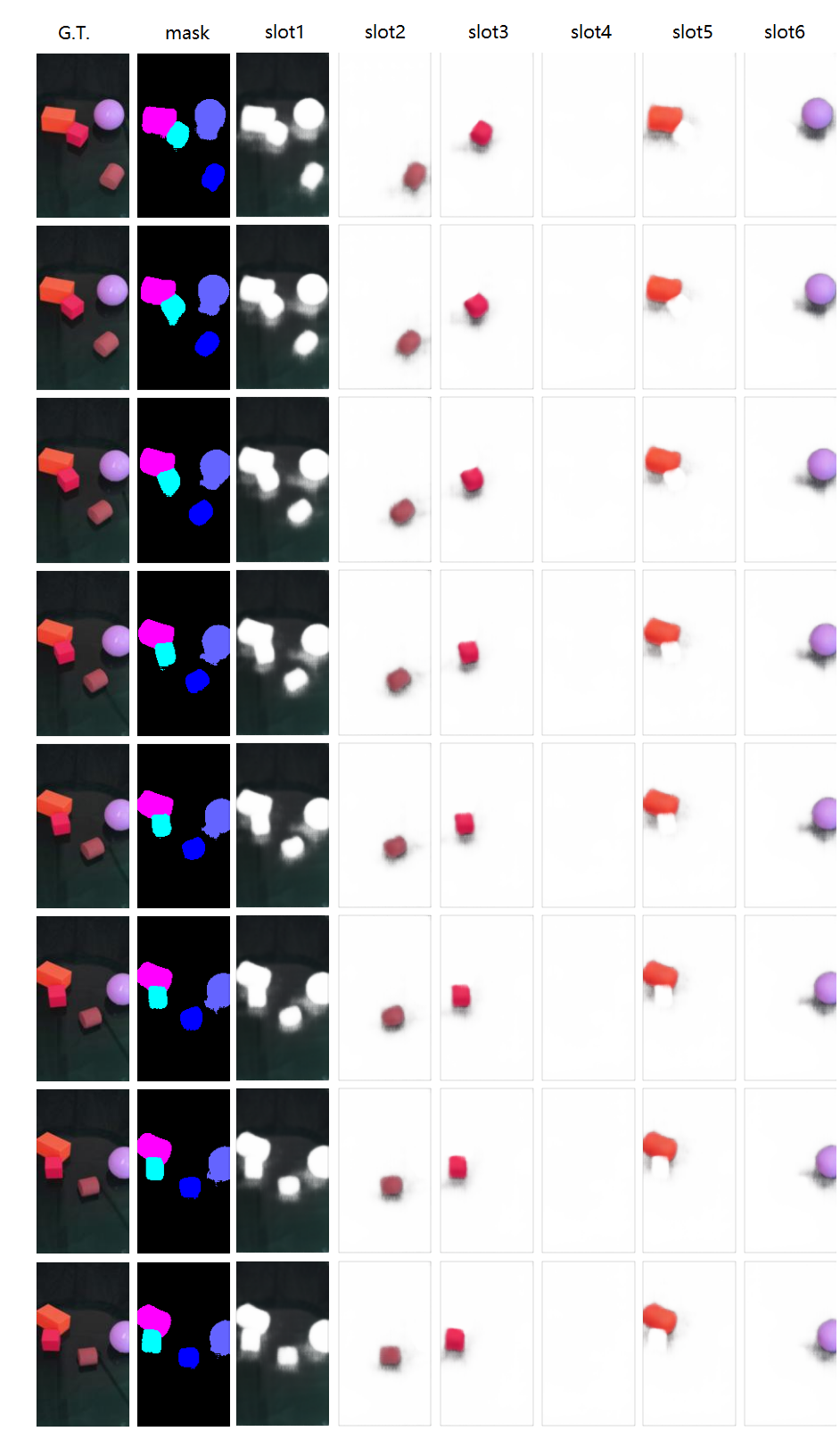}
    \end{minipage}\hfill
    \begin{minipage}{0.48\textwidth}
        \includegraphics[width=\linewidth]{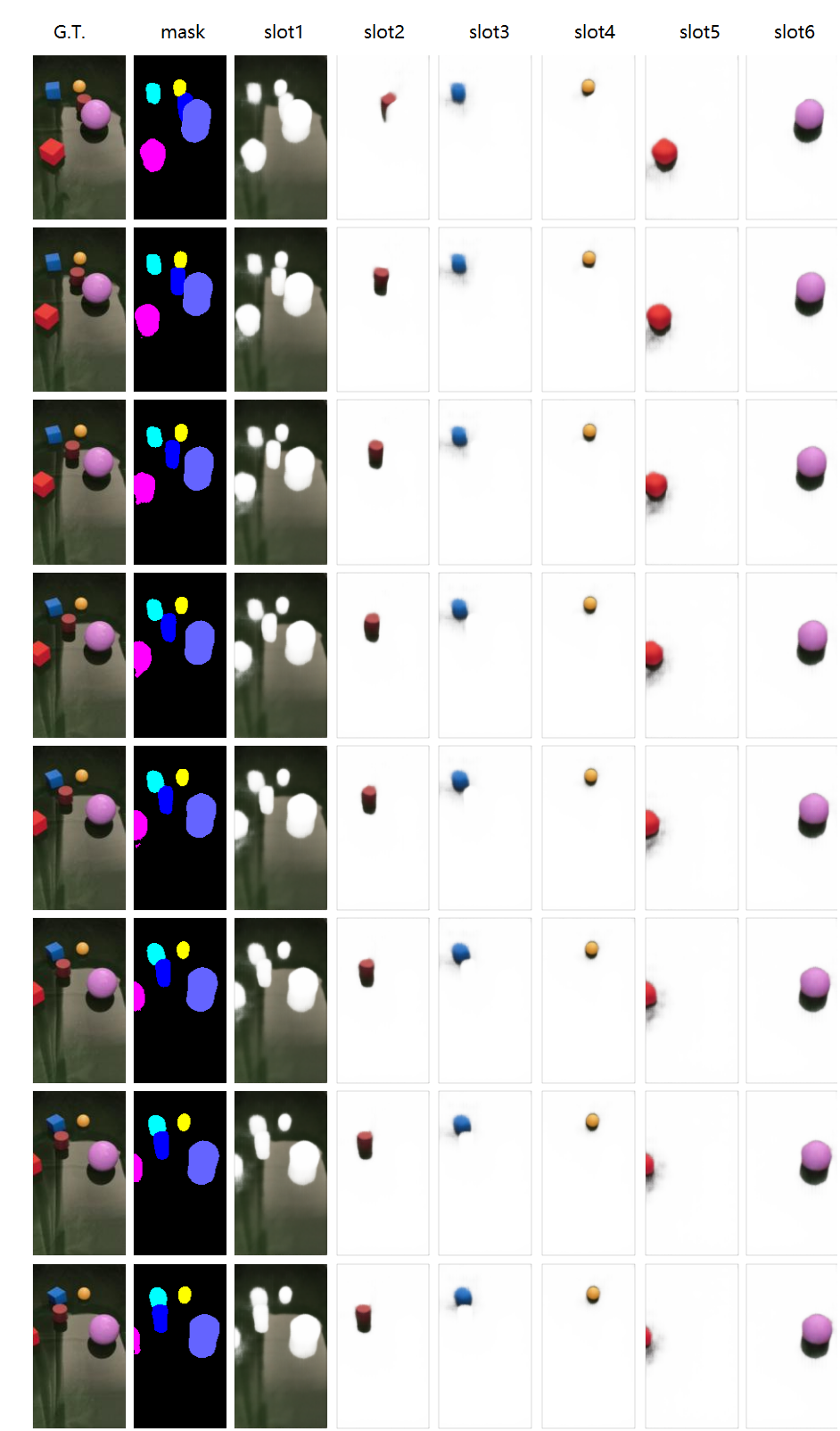} 
    \end{minipage}
    \caption{Additional quality results of scene decomposition on real-
world dataset over time.} 
    \label{fig-s4}
\end{figure*}

\textbf{Additional Real-world Dataset Results.} Table~\ref{tab-s2} presents the evaluation results of scene decomposition using STATM-SAVi on the real-world dataset. We can observe that STATM-SAVi performs well in decomposing the scene. Figure~\ref{fig-s4} provides more qualitative results of scene decomposition. It is evident that, even in challenging scenarios such as severe reflections in fluids or reduced fluid transparency, the model effectively performs the scene decomposition task, with the slots remaining well-organized and not mixing.

\section{Additional Ablation Study}

\begin{table}[t!]
\caption{Evaluations of using $N$ attention blocks within the spatiotemporal reasoning module on the OBJ3D dataset.}
\label{tab-s3}
\vspace{-3pt}
\begin{tabular}{lccc}
\toprule
{\bf Model}&\small {\bf PSNR}&\small {\bf SSIM}&\small {\bf LPIPS$_{\times 100}$}$\downarrow$\\
\midrule
SlotPi (N=1)&32.58&\textbf{0.91}&7.77\\
SlotPi (N=2)&\textbf{32.67}&\textbf{0.91}&\textbf{7.66}\\
SlotPi (N=3)&32.54&\textbf{0.91}&7.91\\
SlotPi (N=4)&32.62&\textbf{0.91}&7.82\\
\bottomrule
\end{tabular}
\vspace{6pt}
\end{table}

\begin{table}[t!]
\caption{Evaluations of different embedding methods on CLEVRER. \textquoteleft MLP\textquoteright\ refers to the method of using an MLP to compute generalized momentum P, coordinates Q, and scene energy. \textquoteleft Attention\textquoteright\ denotes the application of attention mechanisms to this task. \textquoteleft N \textquoteright\ indicates the number of layers in the network structure. }
\label{tab-s4}
\vspace{-3pt}
\begin{tabular}{lcccccc}
\toprule
{\bf Method}&\small {\bf $N_P$}&\small {\bf $N_Q$}&\small {\bf $N_H$}&\small {\bf FG-ARI}\textsf{(\%)}&\small {\bf FG-mIoU}\textsf{(\%)}&\small {\bf ARI}\textsf{(\%)}\\
\midrule
Attention&1&1&1&\textbf{66.12}&\textbf{50.32}&\textbf{64.16}\\
MLP&3&3&3&63.87&48.88&62.59\\
MLP&7&7&7&64.54&49.62&63.41\\
MLP&9&9&9&65.08&49.79&63.61\\
MLP&11&11&11&64.38&49.08&63.13\\
\bottomrule
\end{tabular}
\vspace{0pt}
\end{table}

\textbf{Spatiotemporal reasoning module}. Table~\ref{tab-s3} presents the evaluation results of using $N$ attention blocks within the spatiotemporal reasoning module on the OBJ3D dataset. We observe that when N=1, the model's expressive capacity is significantly insufficient; however, when N=2, the model achieves optimal performance. When N increases to 3 or more, the model begins to exhibit overfitting, and performance starts to decline. Therefore, on the OBJ3D dataset, we opt for a spatiotemporal reasoning module with 2 attention blocks.

\vspace{3pt}

\textbf{Embedding methods}. From Table~\ref{tab-s4}, it is evident that using attention mechanisms for physical embedding significantly outperforms the MLP method. The MLP approach requires 7 or 9 layers to adequately compute the scene's generalized momentum, coordinates, and energy, yet the results are still inferior to those obtained using attention mechanisms. Additionally, employing multiple MLP layers leads to a sharp increase in model parameters.

\end{document}